%% file: arXiv.tex
\documentclass[10pt,twocolumn,letterpaper]{article}

\usepackage{cvpr}               % To produce the CAMERA-READY version

\usepackage{graphicx}
\usepackage{capt-of}
\usepackage{xcolor}
\usepackage{colortbl}
\usepackage{amsmath,amssymb,amsfonts,dsfont,pifont,bm,bbm,mathrsfs,mathtools,nicefrac}
\usepackage{algorithm,algpseudocode,listings}
\usepackage{booktabs,multirow,adjustbox,diagbox,threeparttable}
\definecolor{citeblue}{HTML}{0071bc}
\definecolor{gbypink}{rgb}{0.99, 0.91, 0.95} 
\usepackage[pagebackref,breaklinks,colorlinks,citecolor=citeblue,bookmarks=false]{hyperref}
\usepackage[capitalize]{cleveref}  % Should be loaded after 'hyperref', and works perfectly with 'subfigure'.
\crefname{section}{Sec.}{Secs.}
\Crefname{section}{Section}{Sections}
\crefname{table}{Tab.}{Tabs.}
\Crefname{table}{Table}{Tables}
\crefname{figure}{Fig.}{Figs.}
\Crefname{figure}{Figure}{Figures}
\crefname{equation}{Eq.}{Eqs.}
\Crefname{equation}{Equation}{Equations}
\newcommand{\tocite}[1]{\textcolor{red}{[TO CITE]}}
\newcommand{\method}{BLV\xspace}

\definecolor{upcolor}{RGB}{57,182,74}
\newcommand{\up}[1]{\textcolor{upcolor}{\ $\uparrow$ #1}}

\def\cvprPaperID{2534}
\def\confName{CVPR}
\def\confYear{2023}

\author{
    Yuchao Wang$^{1\dag}$ \quad
    Jingjing Fei$^{2}$ \quad
    Haochen Wang$^{3\dag}$ \quad
    Wei Li$^{2}$ \\
    Tianpeng Bao$^{2}$ \quad
    Liwei Wu$^2$ \quad
    Rui Zhao$^{1,2\ddagger}$ \quad
    Yujun Shen$^{4}$\\[5pt]
    $^1$Shanghai Jiao Tong University \quad
    $^2$SenseTime Research \\
    $^3$Institute of Automation, Chinese Academy of Sciences \quad
    $^4$CUHK \\
    \small{\texttt{ycw991216@163.com}} \quad
    \small{\texttt{shenyujun0302@gmail.com}} \quad  
    \small{\texttt{wanghaochen2022@ia.ac.cn}} \\
    \small{\texttt{\{feijingjing1, liwei1, baotianpeng, wuliwei, zhaorui\}@sensetime.com}} 
}

\begin{document}

\title{Balancing Logit Variation for Long-tailed Semantic Segmentation}
\maketitle

\input{sections/0.abs.tex}
\input{sections/1.intro.tex}

\input{sections/2.related.tex}
\input{sections/3.method.tex}

\input{sections/4.exp.tex}
\input{sections/5.conclusion.tex}
\input{sections/6.ref.tex}

\appendix
% \renewcommand\thefigure{A\arabic{figure}}
% \renewcommand\thetable{A\arabic{table}}  
% \renewcommand\theequation{A\arabic{equation}}
% \setcounter{equation}{0}
% \setcounter{table}{0}
% \setcounter{figure}{0}

% \newpage

\section*{Appendix}
\maketitle
\input{sections/7.supple.tex}
\end{document}

%% file: sections/0.abs.tex
\begin{abstract}

Semantic segmentation usually suffers from a long-tail data distribution.  % , which could cause biased predictions.
Due to the imbalanced number of samples across categories, the features of those tail classes may get squeezed into a narrow area in the feature space.  % , leaving the decision boundaries hard to learn.
Towards a balanced feature distribution, we introduce category-wise variation into the network predictions in the training phase such that an instance is no longer projected to a feature point, but a small region instead.
Such a perturbation is highly dependent on the category scale, which appears as assigning smaller variation to head classes and larger variation to tail classes.
In this way, we manage to close the gap between the feature areas of different categories, resulting in a more balanced representation.  % and hence alleviate the learning difficulty.
It is noteworthy that the introduced variation is discarded at the inference stage to facilitate a confident prediction.
Although with an embarrassingly simple implementation, our method manifests itself in strong generalizability to various datasets and task settings.
Extensive experiments suggest that our plug-in design lends itself well to a range of state-of-the-art approaches and boosts the performance on top of them.%
\footnote{
Code: \url{https://github.com/grantword8/BLV}.

~$^\dag$This work was done during the internship at SenseTime Research.

~$^\ddagger$Rui Zhao is also with Qing Yuan Research Institute, Shanghai Jiao Tong University.
}

\iffalse
%
Due to the long-tailed distribution of training set, semantic segmentation model is severely biased towards the dominant category. 
%
We observe that vanilla training on long-tailed data with cross-entropy loss makes the instance-rich head classes severely squeeze the spatial distribution of the tail classes, which leads to insufficient training of tail class pixels. 
%
It is unfavorable for training on balanced data, but can be utilized to adjust the validity of the samples in long-tailed data, thereby solving the distorted embedding
space of long-tailed traning data. 
%
To this end, this paper proposes the a novel loss called "Augment Your Logits"(\method) to augment the final logits output of the training model.
%
We set relatively large the amplitude of perturbation to the tail class logits and small that of the head class.
%
The large cloud size can reduce the softmax saturation and thereby making tail class samples more active as well as enlarging the embedding space. 
%
Extensive experiments on benchmark datasets and three main stream semantic segmentation tasks validate the superior performance of the proposed method. 
%
Source code will be made public available.
%
\fi

\end{abstract}

%% file: sections/1.intro.tex
\section{Introduction}\label{sec:intro}

The success of deep models in semantic segmentation~\cite{fcn, unet, pspnet, deeplab} benefits from large-scale datasets.
However, popular datasets for segmentation, such as PASCAL VOC~\cite{voc} and Cityscapes~\cite{cityscapes}, usually follow a long-tail distribution, where some categories may have far fewer samples than others.
Considering the particularity of this task, which targets assigning labels to pixels instead of images, it is quite difficult to balance the distribution from the aspect of data collection.
Taking the scenario of autonomous driving as an example, a bike is typically tied to a smaller image region (\textit{i.e.}, fewer pixels) than a car, and trains appear more rarely than pedestrians in a city.
Therefore, learning a decent model from long tail data distributions becomes critical.

% Deep learning has made incredible progress in various computer vision tasks, such as image classification, object detection and semantic segmentation.
% %
% The keys to this success are the large-scale datasets, high-performance GPUs and advancement of neural network architectures.
% %
% In real-world applications, training data typically follows a long-tailed categorical distribution, where some categories are extremely numerous and others are extremely few.
% %
% This imbalance data distribution leads to challenging model training by empirical risk minimization, making the trained model biased towards head category and causing poor performance on tail category.
% %
% Semantic segmentation, as one of the most instance-intensive task, suffers badly from the undesirable consequences of long-tailed distributions.

\definecolor{myyellow}{RGB}{184,146,48}
\definecolor{myblue}{RGB}{0,0,255}
\begin{figure}[t]
    \centering
    \includegraphics[width=1.0\linewidth]{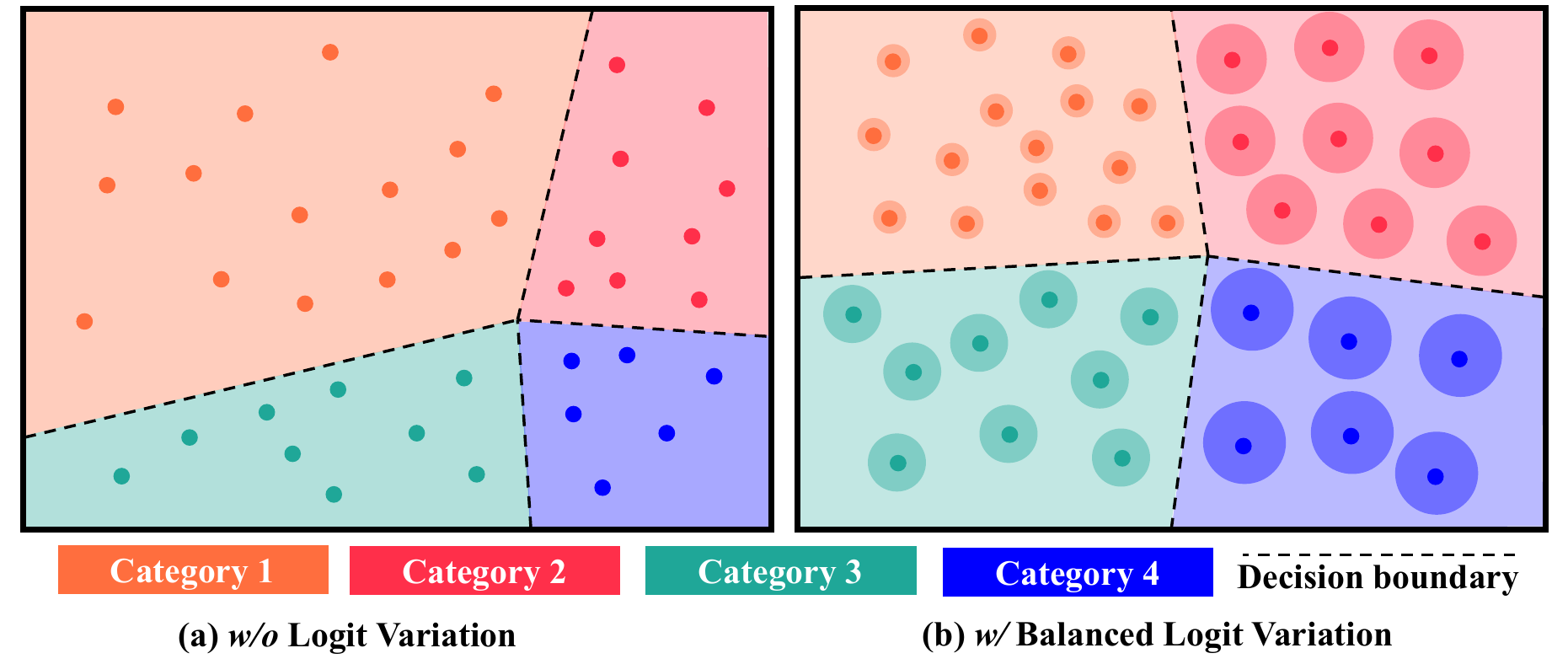}
    \vspace{-15pt}
    \caption{%
        \textbf{Illustration of logit variation} from the feature space, where each point corresponds to an instance and different colors stand for different categories.
        (a) Without logit variation, the features of tail classes (\textit{e.g.}, the \textbf{\textcolor{myblue}{blue}} one) may get squeezed into a narrow area.
        (b) After introducing logit variation, which is controlled by the category scale (\textit{i.e.}, number of training samples belonging to a particular category), we expand each feature point to a feature region with random perturbation, resulting in a more category-balanced feature distribution.
    }
    \label{fig:stats}
    \vspace{-10pt}
\end{figure}

A common practice to address such a challenge is to make better use of the limited samples from tail classes.
For this purpose, previous attempts either balance the sample quantity (\textit{e.g.}, oversample the tail classes when organizing the training batch)~\cite{chawla2002smote, shen2016relay,galar2013eusboost,hao2020annealing,kim2020m2m}, or balance the per-sample importance (\textit{e.g.}, assign the training penalties regarding tail classes with higher loss weights)~\cite{li2021autobalance,park2021influence,cui2019class,ren2018learning}.
Given existing advanced techniques, however, performance degradation can still be observed in those tail categories.

This works provides a new perspective on improving long-tailed semantic segmentation.
Recall that, in modern pipelines based on neural networks~\cite{fcn, deeplab, xie2021segformer, zhang2021k, zheng2021rethinking, chen2022vision}, instances are projected to representative features before categorized to a certain class.
We argue that the features of tail classes may get squeezed into a narrow area in the feature space, as the blue region shown in \cref{fig:stats}a, because of miserly samples.
To balance the feature distribution, we propose a simple yet effective approach via introducing \textbf{balancing logit variation (BLV)} into the network predictions.
Concretely, we perturb each predicted logit with a randomly sampled noise during training.
That way, each instance can be seen as projected to a feature region, as shown in \cref{fig:stats}b, whose radius is dependent on the noise variance.
We then propose to balance the variation by applying smaller variance to head classes and larger variance to tail classes so as to close the feature area gap between different categories.
This newly introduced variation can be viewed as a special augmentation and discarded in the inference phase to ensure a reliable prediction.

We evaluate our approach on three different settings of semantic segmentation, including fully supervised~\cite{yuan2020object, xiao2018unified, dosovitskiy2020image, pspnet}, semi-supervised~\cite{zou2020pseudoseg, chen2021semi, wang2022semi, zhong2021pixel}, and unsupervised domain adaptation~\cite{du2022learning, li2022class, zhang2021prototypical, zhou2022uncertainty}, where we improve the baselines consistently.
We further show that our method works well with various state-of-the-art frameworks~\cite{hoyer2022daformer, hoyer2022hrda, xie2021segformer, zhang2021k, amini2022self, wang2022semi} and boosts their performance, demonstrating its strong generalizability.

%% file: sections/2.related.tex
\section{Related Work}\label{sec:related-work}

\noindent\textbf{Semantic segmentation.}
%
% Prior arts usually improve the performance of this task via either designing more effective model architectures or constructing better representations.
%
Network architecture for semantic segmentation has evolved for years, from CNNs~\cite{fcn, pspnet, deeplab} to Transformers~\cite{zhang2021k, zheng2021rethinking, xie2021segformer, chen2022vision, dosovitskiy2020image}.
%
% At the beginning of deep learning era, CNN-based architectures~\cite{fcn,pspnet,deeplab} like FCN~\cite{fcn} have become a mainstream framework.
%
% More recent Transformer-based~\cite{zhang2021k, zheng2021rethinking, xie2021segformer, chen2022vision, dosovitskiy2020image} architecture have proved higher performance for semantic segmentation due to its effectiveness.
%
Another line of research works focuses on enhancing the extracted representations like integrating attention mechanisms~\cite{fu2019dual, huang2019ccnet, li2018pyramid, zhong2020squeeze} or context representations~\cite{yuan2020object, wang2020deep, zhang2018context, lin2018multi, yu2020context, yuan2021ocnet} into segmentation models.
\method is complementary to these various frameworks and improves several state-of-the-art methods consistently.

\noindent\textbf{Semi-supervised semantic segmentation.}
To alleviate the heavy need for large-scale annotated data, semi-supervised semantic segmentation has become a research hotspot.
There are two typical frameworks for this task: consistency regularization~\cite{em, chen2021semisupervised, french2019semi} and self-training~\cite{fixmatch, dash, cct, bachman2014learning}.
Consistency regularization applies various perturbations~\cite{french2019semi} on training data and forces consistent predictions between the perturbed and the unperturbed input~\cite{em}.
Self-training~\cite{zuo2021self, st++, wang2022semi, xie2020self, cct, cutmix, ael} uses the predictions from the pre-trained model as the ``ground-truth'' of the unlabeled data and then trains a semantic segmentation model in a fully-supervised manner.
These two frameworks have no specialized operations for long-tail data.
To this end, we provide a concise and generic approach that can be integrated into any framework.

\noindent\textbf{Unsupervised domain adaptive semantic segmentation.}
UDA semantic segmentation aims at learning segmentation model that transfer knowledge from labeled source domain to unlabeled target domain.
Early methods for UDA segmentation focus on enabling the model to extract domain-invariant features.
They align the cross-domain feature distribution at image level~\cite{hoffman2018cycada, sankaranarayanan2018learning, gong2019dlow}, feature level~\cite{chang2019all, chen2019progressive, li2021semantic, tsai2018learning} and output level~\cite{melas2021pixmatch, vu2019advent, tsai2018learning} via image style transfer~\cite{yang2020fda, guo2021label, hoffman2018cycada, li2019bidirectional, kim2020learning}, image feature domain discriminator~\cite{tsai2019domain, gan, nowozin2016f, pan2020unsupervised, wang2020differential} or well-designed metrics~\cite{guo2021metacorrection, long2015learning, lee2019sliced}.
Follow-up study~\cite{zhang2021prototypical, chen2022deliberated} suggests that the self-training-based pipeline leads to more consistent improvement.
Recently, DAFormer~\cite{hoyer2022daformer} and HRDA~\cite{hoyer2022hrda} provide a self-training-based Transformer architecture together with many efficient training strategies, which can achieve consistent improvement over other competitors.
\method can be simply integrated into existing pipelines, and consistently improve their performance.

\noindent\textbf{Long-tail learning.} Since the long-tail phenomenon is common~\cite{yang2022survey} in deep learning, the performance of the model tends to be dominated by the head category, while the learning of the tail category is severely underdeveloped.
One intuitive solution to alleviate unbalanced data distribution is data processing, which typically consists of three ways: over-sampling~\cite{haixiang2017learning, wu2020forest, peng2020large, gupta2019lvis}, under-sampling~\cite{haixiang2017learning, buda2018systematic, sinha2020class, tan2020equalization} and data augmentation~\cite{zang2021fasa, chu2020feature, liu2022breadcrumbs, chou2020remix}. Various methods have been proposed to alleviate the long-tail phenomenon in semantic segmentation, which can be mainly divided into three settings: fully supervised~\cite{tian2022striking,berman2018lovasz}, semi-supervised~\cite{he2021re, hu2021semi,fan2022ucc}, and UDA~\cite{ruan2019category,zou2018unsupervised,yang2020adversarial,li2022class}. It is noteworthy that existing methods are usually limited to a specific setting and lack generalizability.

\noindent\textbf{Noise-based augmentation.}
To improve model robustness and avoid over-fitting, augmenting data with noise~\cite{holmstrom1992using, bengio2011deep, ding2016convolutional} at image level or feature level is widely applied to model training.
Techniques~\cite{lopes2019improving,zur2009noise} like Dropout~\cite{srivastava2014dropout}, color jittering~\cite{afifi2019else}, gaussian noise, are the most common methods and proved to be simple yet efficient, but they might also introducing task-agnostic bias~\cite{yin2015noisy}.
Besides, methods like \textit{M2m}~\cite{kim2020m2m} and \textit{AdvProp}~\cite{xie2020adversarial} utilize adversarial examples to augment the training data and significantly improve model robustness.
Prior arts focus on improving the robustness yet ignoring the prevalence of long-tail data,
whereas our \method can alleviate the feature squeeze caused by long-tail data effectively.
Logit adjustment~\cite{menon2021long} has become a popular strategy to alleviate the long-tail issue and hence owns numerous variants.
GCL~\cite{li2022long} proposed a two-stage logit adjustment method that involves perturbating features with Gaussian noise and re-sampling classifier learning which has demonstrated promising results in long-tailed classification tasks.
As another variant of logit adjustment, we apply it to the long-tailed segmentation task. Through single-stage training only, \method enhances baseline performance in fully supervised, semi-supervised, and domain adaptative settings. Notably, \method has a robustness that allows it to manage non-Gaussian adjustment terms and variations in the adjustment term.
% GCL~\cite{li2022long}, similar to our \method, adjusts the logit to alleviate softmax saturation and retrains the classifier using the strategy of effective sample number sampling.
% %
% Both \method and GCL~\cite{li2022long} share the same goal of alleviating the imbalance in feature space distribution.
% %
% Nonetheless, the implementation of \method is simpler and mainly focused on segmentation tasks.
% %
% Furthermore, \method's investigations extend to estimating the distribution of unlabeled data, notably in the semi-supervised and unsupervised domain adaptive settings, where the category distribution of training data is unknown.

%% file: sections/3.method.tex
\section{Method}\label{sec:method}

In this section, we first formulate our problem mathematically and elaborate our approach detailedly in \cref{sec:method_1}.
Then we specify how \method can be used in three tasks where the settings are not exactly the same, \textit{i.e.}, fully-supervised, semi-supervised, domain adaptive settings, in \cref{sec:method_2}, \cref{sec:method_3}, \cref{sec:method_4}, respectively.

\subsection{Elaboration of \method}\label{sec:method_1}
Long-tailed label distribution is detrimental to the training of deep learning models.
As \cref{fig:stats}a illustrated the total numbers of instances from tail categories are extremely much fewer when compared to head categories.
As a result, they are squeezed into a very small area in the feature space,
which means \textit{the decision boundaries of these tailed categories can be severely biased}.
Thus, at the inference stage, many similar data outside the distribution of the training tail category instances will be misclassified due to this squeeze.
Next, we will elaborate on our approach.

Given a long-tailed training dataset with $N$ labeled images of $C$ categories: $D=\left\{(x_{image}^{i}, y_{image}^{i})\right\}_{i=1}^{N}$, where $y_{image}^{i}\in \left\{0, 1, ..., C-1\right\}$,  
our goal is to train a semantic segmentation model $f_{model}$ with more balanced representations.
To achieve this, we need to take a more fine-grained perspective.
For segmentation tasks, the corresponding task-related instances are pixels, instead of images.
Thus we can view the task as a multi-label classification task at the pixel level.
During the training stage, assuming there is an input data batch $X_{batch}$ with a shape of $\left\{B, 3, H, W\right\}$ and its corresponding label $Y_{batch}$, where $B$ is the batch size and $H, W$ denotes the size of the images, we can input it into the model $f$ to get an output vector $\tilde{Z}_{batch}$.
The shape of $\tilde{Z}_{batch}$ will be $\left\{B, C, H, W\right\}$ (we assume that $H, W$ remain the same here for simplicity because $\tilde{Z}_{batch}$ can be upsampled to this size), where $C$ is the number of categories.

From the view of instances (\textit{i.e.}, pixels in segmentation task), we can reshape the output $\tilde{Z}_{batch}$ from $\left\{B, C, H, W\right\}$ into $\left\{B\times H \times W, C\right\}$.
So for this batch, we have $B\times H \times W$ pixels and corresponding $C$-dimensional prediction for each of them.
Taking pixel $i$ as an example, its output $\tilde{Z}_{batch}^{i} = \left[z_{0}^{i}, z_{1}^{i}, \cdots, z_{C-1}^{i}\right]$.
In order to calculate the cross entropy loss during training, we need to convert it into probabilities by the \texttt{softmax} formula \cref{eq:softmax}.
\begin{equation}\label{eq:softmax}
     p^{i}_{k} = \frac{e^{z^{i}_{k}}}{\sum_{j=0}^{C-1}e^{z^{i}_{j}}},
\end{equation}
where $p^{i}_{k}$ denotes the probability of pixel $i$ to be of category $k$ and $C$ is the number of categories.
After obtaining the probabilities, common practices are to use them to calculate the Cross-Entropy Loss in \cref{eq:cross_entropy}.
\begin{equation}\label{eq:cross_entropy}
    L_{CE}(\tilde{Z}_{batch}^{i}) = -\sum_{k=0}^{C-1} y_{k}^{i}\log{p^{i}_{k}},
\end{equation}
where $y_{k}^{i}$ is $k$-th term of the one-hot encoded ground truth $\left[y_{0}^{i}, y_{1}^{i}, \cdots, y_{C-1}^{i}\right]$.
\begin{figure}[t]
    \centering
    \includegraphics[width=1.0\linewidth]{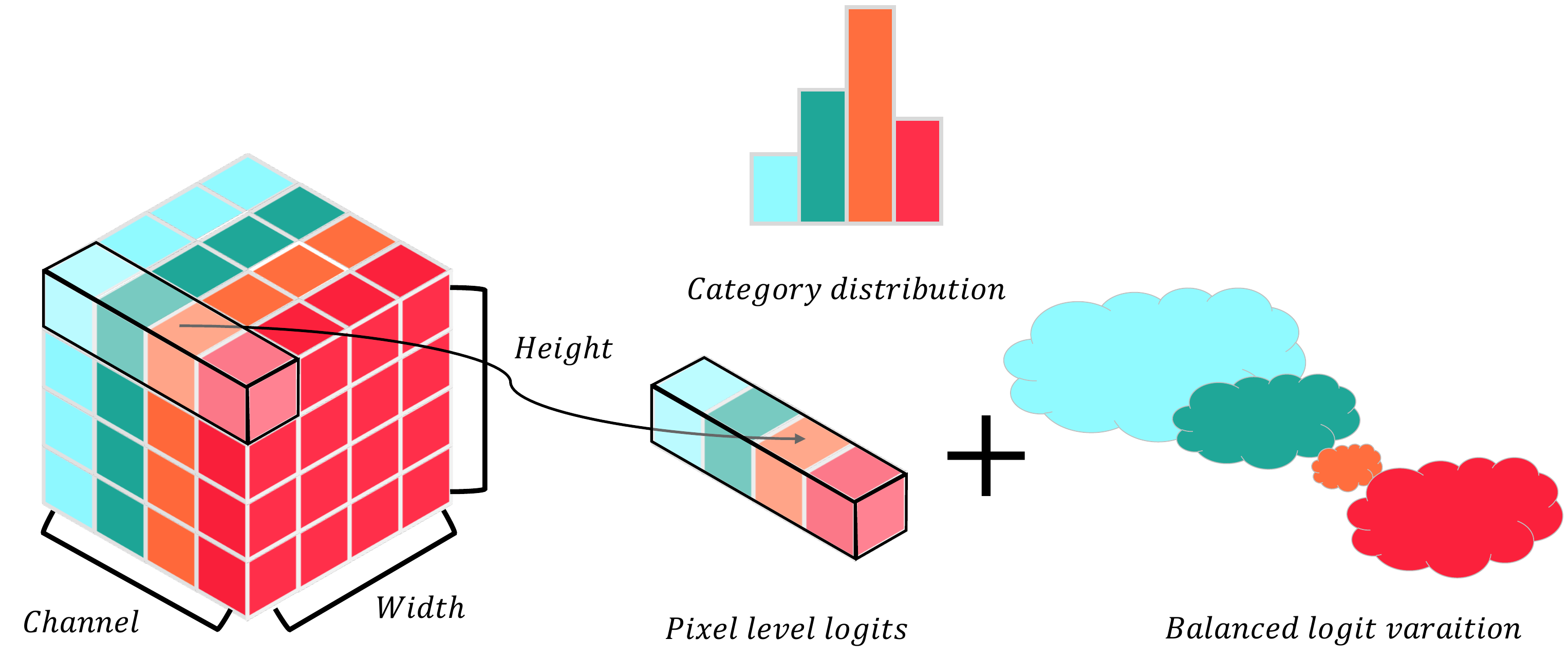}
    \vspace{-15pt}
    \caption{%
        Diagram of the introduction of balanced logit variation, where we perturb the per-pixel logit with a category-specific noise.
        The noise variance is in inverse proportion to the category scale.
    }
    \label{fig:method}
    \vspace{-10pt}
\end{figure}
Every $z_{k}^{i}$ is defined as the \textbf{logit} for the instance (\textit{i.e.}, pixel) $i$.
The step-by-step derivation from \cref{eq:softmax} to \cref{eq:cross_entropy} depicts a direct relationship between the $L_{CE}$ optimization and logit term $z$.
Logit term $z$ is critical to the long-tail problem.
Because the dimensionality of logit $z$ is consistent with the total number of categories and directly affects the computation of the loss, making it the most intuitive way to affect the size of the categorical area feature space.
Then the crux lies in how to use logit to alleviate the long-tail problem.
One intuitive way is simply to rescale the logit according to the category frequency~\cite{menon2021long}.
However, semantic segmentation is an extremely instances-intensive task, so simply rescaling the logit fixedly according to the category frequency leads to overfitting problems.

To this end, we propose to add variation into the network predictions (\textit{i.e.} $z$ here) in \cref{eq:balancing}.
\begin{equation}\label{eq:balancing}
    \hat{z}^{i}_{k} = z^{i}_{k} + \frac{c_{k}}{\max_{i=0}^{C-1}c_{i}}|\delta(\sigma)|, \quad c_{k} =\log \frac{\sum_{j=0}^{C-1}q_{j}}{q_{k}}
\end{equation}
where $q_{k}$ is the number of the instances with category $k$ and $\delta$ is a gaussian distribution with a mean of \textbf{$0$} and standard deviation of $\sigma$.
\cref{eq:balancing} is quite easy to understand, as it assigns smaller variation to the head categories and larger variation to the tail categories.
By adding this variation, which is inversely proportional to the category scale, our method can be equivalent to \textit{expanding the distribution of each instance over the feature space from a single point into a small region}.
Therefore, when training under this setting, we can obtain a more category-balanced feature representation space.
We give a more straightforward explanation of our approach in \cref{fig:method}.
Besides, the only hyper-parameter of \cref{eq:balancing} is the $\sigma$, making it easy to generalize to other tasks.
The form of variation can actually be not limited to Gaussian distribution, in the ablation experiment \cref{sec:ablation} we found that variation sampled from other distributions can also work.
It is noteworthy that the introduced variation is discarded at the inference stage to facilitate a confident prediction.
Next, we will elaborate on how to specifically apply \cref{eq:balancing} for different settings of long-tail semantic segmentation tasks.

\subsection{\method for Fully Supervised Segmentation}\label{sec:method_2}
Since the labels of the training data are available in the fully supervised semantic segmentation task, the category-by-category distribution can be obtained easily when preprocessing the data.
It should be noted that since the instances of the segmentation task are pixels, the number of pixels of each class needs to be counted before obtaining their distribution.
We present all experimental results for this task in \cref{sec:fully_super}.
\subsection{\method for Semi-Supervised Segmentation}\label{sec:method_3}
Semi-supervised semantic segmentation is a more challenging task, due to the fact that only a small portion of the training images are carefully labeled~\cite{zhang2020survey}.
A simple approach is to equate the pixel-level category distribution of the labeled images with the category distribution of the whole training set (including both the labeled and the unlabeled images).
This estimated distribution is quite inaccurate when the labeled/unlabeled division is very extremely unbalanced, for example, the $1/16 (186)$ partition protocols in \cref{sec:semi_exp}. 

Therefore, we propose an epoch-based update strategy of the distribution to make it closer to the true distribution.
Suppose after $n$ epochs of training, we have a model $f_{n}$. 
For all the unlabeled training images set $X^{u}_{image}$, we infer the labels of all the images in $X^{u}_{image}$ thus get its corresponding pseudo-label: $\hat{Y}^{u}_{image}$.
Thus, we calculate the number of pixels in each category by the following formula \cref{eq:cal_distribution}.
\begin{equation}\label{eq:cal_distribution}
    q_{k} = \frac{\sum_{n=1}^{N}\sum_{m=1}^{H\times W}\mathbbm{1}\left[\hat{y}_{nm}^u = k\right]}{\sum_{i=0}^{C-1}\sum_{n=1}^{N}\sum_{m=1}^{H\times W}\mathbbm{1}\left[\hat{y}_{nm}^u = i\right]}
\end{equation}
where $q_{k}$ denotes the $k$-th category frequency, $\hat{y}_{nm}^u$ denotes the $m$-th element of the $n$-th pseudo-label and $N$ is the number of the unlabeled images.

\cref{eq:cal_distribution} will be used to calculate the updated category distribution $\left\{q_{0}, q_{1},\cdots,q_{n-1}\right\}$ after every epoch.
Then we can bring this estimated distribution into \cref{eq:balancing}.
With this design, our approach can efficiently and consistently improve the performance of semi-supervised semantic segmentation tasks.

\subsection{\method for UDA Segmentation}\label{sec:method_4}
Unsupervised domain adaptive semantic segmentation attempts to train a model that works well on the target domain by using labeled source domain data and unlabeled target domain data.
Since the goal is to improve the performance of the model on the target domain, yet the images on this domain are unlabeled, this poses a tricky problem.
A widely recognized perspective is to view UDA semantic segmentation as semi-supervised semantic segmentation~\cite{zhang2021prototypical}.
Because the source domain data is naturally labeled, this perspective makes certain sense without considering the inter-domain gap.
Therefore, we can estimate the distribution of the target domain data with \cref{eq:cal_distribution}.
However, for the UDA semantic segmentation, we propose a more concise way: viewing the category distribution of the source domain data as if it were the category distribution of the target domain.
The data from the source domain are typical computer-rendered synthetic labeled images while the data from the target domain are generally a real-world collection of images.
This difference makes the images of the two domains significantly different only in terms of style, and essentially identical in terms of contextual relationships and category distribution.
In \cref{exp:uda} we have experimentally demonstrated that this simple estimation method works.

%% file: sections/4.exp.tex
\section{Experiments}\label{sec:exp}
We present experimental results on three main-stream segmentation tasks: semantic segmentation, semi-supervised semantic segmentation, and domain adaptive semantic segmentation.
Besides, we present comparisons with previous works towards class-imbalanced problems.
The mean of Intersection over Union (mIoU) is adopted as the metric to evaluate all the results.

\subsection{Towards Fully Supervised Setting}\label{sec:fully_super}
\noindent\textbf{Datasets.} We used the typical long-tailed dataset: Cityscapes~\cite{cityscapes}.
Cityscapes is a driving dataset for semantic segmentation, which consists of 5000 high-resolution images for training and 500 images for validation.
We first resize the training images into a resolution of $2048\times1024$, then crop them into $512\times1024$.

\noindent\textbf{Implementation details.} 
We used the mmsegmentation codebase~\cite{mmseg2020} and trained all the models with 8 Tesla V100 GPUs.
To validate the proposed method \method, we apply our method \method to various state-of-the-art semantic segmentation models including ResNet~\cite{he2016deep}, Swin-Transformer~\cite{liu2021swin}, Mix-Transformer~\cite{xie2021segformer}, ViT~\cite{dosovitskiy2020image} based encoder with OCRHead~\cite{yuan2020object}, K-NeT~\cite{zhang2021k}, PSPHead~\cite{pspnet}, SegformerHead~\cite{xie2021segformer}, UperHead~\cite{xiao2018unified} based decoders respectively.
The batch size is set to 16 for all models.
The training iterations are $160k$ for MiT-b0 + SegformerHead, $80k$ for Swin-T + K-Net and Vit-B16 + UperHead, $40k$ for all the other models.
We use AdamW optimizer for three transformer-based models: Swin-T + K-Net, MiT-b0 + SegformerHead, and Vit-B16 + UperHead, with a learning rate of $6\times10^{-5}$, weight decay of $0.01$, a linear learning rate warmup with $1.5k$ iterations and linear decay afterwards.
For all of the other models, we use the same configuration: SGD optimizer with a learning rate of $0.01$, a weight decay of $5\times 10^{-4}$.

\noindent\textbf{Results.} 
Table \cref{tab:fully_supervised} summarizes the detailed comparison results across different architectures.
We observe that our method boosts all of these baseline models consistently.
Equipped with our method, these models gains $+0.72\%$, $+0.43\%$, $+0.55\%$, $+0.24\%$, $+0.47\%$, $+0.35\%$, $+1.20\%$ respectively without any additional model parameters.
The performance gains on various models with different network structures, including CNN-based and Transformer based models, indicate that our method is universal and can be generalized to various segmentation models.
To verify the effectiveness of \method towards long-tail data, we compute mIoU on \textbf{9} tail categories: \textit{Wall}, \textit{T.light}, \textit{Sign}, \textit{Rider}, \textit{Truck}, \textit{Bus}, \textit{Train}, \textit{M.bike}, \textit{Bike}.
The ``mIoU (tail)'' column demonstrates that the \method indeed boosts the performance of these tail categories by a large margin.

\subsection{Towards Semi-Supervised Setting}\label{sec:semi_exp}
\noindent\textbf{Datasets.} Semi-Supervised semantic segmentation aims to learn a model with only a few labeled samples.
Two typical benchmark datasets are usually used for validation: PASCAL VOC 2012\cite{voc} and Cityscapes\cite{cityscapes}.
PASCAL VOC 2012 is a class-balanced and simpler dataset.
Therefore, we mainly conduct experimental verification on Cityscapes.
We follow the commonly used $1/16$, $1/8$, $1/4$ and $1/2$ partition protocols, that is, only the corresponding fraction number have labels, and the rest of the images are considered unlabeled.
It is worth mentioning that our method adopts the generally used sliding window evaluation when evaluating.
\noindent\textbf{Implementation details.} The classical Self-Training~\cite{amini2022self} without any other tricks as the baseline due to its simplicity and to be consistent with our proposed method.
The core of semi-supervised segmentation methods is the training strategy, not the network structure.
So we use ResNet-101~\cite{he2016deep} as the backbone and DeepLabv3+~\cite{deeplab} as the decoder.
We use stochastic gradient descent (SGD) optimizer with an initial learning rate of $0.01$, and weight decay as $0.0005$.
The momentum coefficient $\mu$ for Teacher model~\cite{tarvainen2017mean} updating is set to $0.999$.
The crop size is set as $769\times 769$ and batchsize is set as 16.

\noindent\textbf{Results.} With our proposed \method, \cref{tab:semi} demonstrates that naive self-training framework achieves consistent performance gains over the naive self-training baseline by $+1.05\%$, $+1.26\%$, $+1.49\%$, $+0.99\%$ under $1/16$, $1/8$, $1/4$ and $1/2$ partition protocols.
To verify the effectiveness of \method towards long-tail data in semi-supervised segmentation task, we also list the ``mIoU(tail)'' column as the \cref{sec:fully_super}.
This demonstrates that the \method indeed improves the performance of the tail categories.
\begin{table}[t]
\centering
\caption{%
Experiments across architectures for fully semantic segmentation tasks on \textbf{Cityscapes} \textit{validation} set.
The green arrows indicate the relative improvement in performance.
}
\label{tab:fully_supervised}
\vspace{-8pt}
\setlength{\tabcolsep}{3pt}
\scalebox{0.85}{
\begin{tabular}{ll|ll}
\toprule
Backbone & Decoder & mIoU & mIoU (tail) \\
\midrule
\multirow{2}{*}{HRNet-18~\cite{sun2019deep}} & OCRHead~\cite{yuan2020object} & 79.22 & 63.51\\
&\textbf{ + \method} & \textbf{79.94\up{0.72}} & \textbf{66.70\up{3.19}}\\
\midrule 
\multirow{2}{*}{ResNet50~\cite{deeplab}} &UperHead~\cite{xiao2018unified} &78.28&62.56 \\
&\textbf{ + \method} & \textbf{78.63\up{0.35}}& \textbf{64.57\up{2.01}}\\
\midrule
\multirow{2}{*}{ResNet50~\cite{he2016deep}} & PSPHead~\cite{pspnet}& 77.98 & 61.96 \\
 &\textbf{ + \method} & \textbf{78.53\up{0.55}}&\textbf{63.34\up{1.38}}\\
\midrule
\multirow{2}{*}{ResNet101~\cite{he2016deep}} &UperHead~\cite{xiao2018unified} &79.41 &64.68 \\
&\textbf{ + \method} & \textbf{79.88\up{0.47}}&\textbf{66.29\up{1.61}}\\

\midrule
\multirow{2}{*}{MiT-b0~\cite{xie2021segformer}}& SegformerHead~\cite{xie2021segformer} & 76.85&67.58 \\
&\textbf{ + \method} & \textbf{77.09\up{0.24}}&\textbf{68.91\up{1.33}}\\
\midrule
\multirow{2}{*}{Swin-T~\cite{liu2021swin}} & K-NeT~\cite{zhang2021k} & 79.68& 71.70\\
&\textbf{ + \method} & \textbf{80.11\up{0.43}} & \textbf{72.94\up{1.24}}\\
\midrule
\multirow{2}{*}{Vit-B16~\cite{dosovitskiy2020image}}&UperHead~\cite{xiao2018unified}&76.48&68.25\\
&\textbf{ + \method} & \textbf{77.68\up{1.20}}&\textbf{70.63\up{2.38}}\\
\bottomrule
\end{tabular}}
%\vspace{-10pt}
\end{table}

% \begin{table}[t]
% \centering
% \caption{%
% Experiments on semi-supervise semantic segmentation tasks on \textbf{Cityscapes} \textit{validation} set. 
% %
% The green arrows indicate the relative improvement in performance.
% }
% \label{tab:semi}
% \vspace{-8pt}
% \setlength{\tabcolsep}{4pt}

% \begin{tabular}{cccc}
% \toprule
% 1/16 (186) & 1/8 (372)& 1/4 (744) & 1/2 (1488) \\
% \midrule
% \multicolumn{4}{c}{\textit{Self-Training}} \\
% 68.21 & 72.01 & 74.03 &77.99 \\
% \midrule
% \multicolumn{4}{c}{\textit{Self-Training + \method}} \\
% \textbf{69.26\up{1.05}} &\textbf{73.27\up{1.26}} &\textbf{75.52\up{1.49}} & \textbf{78.98\up{0.99}} \\

% \bottomrule
% \end{tabular}
% \vspace{-5pt}
% \end{table}

\begin{table}[t]
\centering
\caption{%
Experiments on semi-supervised semantic segmentation tasks on \textbf{Cityscapes} \textit{validation} set. 
The green arrows indicate the relative improvement in performance.
}
\label{tab:semi}
\vspace{-8pt}
\setlength{\tabcolsep}{10pt}
\scalebox{0.85}{
\begin{tabular}{ll|ll}
\toprule
Partition & Method & mIoU& mIoU (tail)  \\
\midrule
\multirow{2}{*}{1/16 (186)} & Self-Training & 68.21 & 53.09   \\
 &\textbf{+\method} &\textbf{69.26\up{1.05}} &\textbf{55.23\up{2.14}} \\
\midrule
\multirow{2}{*}{1/8 (372)} & Self-Training & 72.01 &58.74   \\
 &\textbf{+\method}&\textbf{73.27\up{1.26}} & \textbf{60.33\up{1.59}}  \\
\midrule
\multirow{2}{*}{1/4 (744)} & Self-Training & 74.03 &  61.76 \\
 &\textbf{+\method}& \textbf{75.52\up{1.49}} & \textbf{63.51\up{1.75}}  \\
\midrule
\multirow{2}{*}{1/2 (1488)} & Self-Training & 77.99 &  65.96 \\
 &\textbf{+\method}& \textbf{78.98\up{0.99}} & \textbf{67.24\up{1.28}}   \\

\bottomrule
\end{tabular}}
\vspace{-10pt}
\end{table}

\subsection{Towards UDA Setting}\label{exp:uda}

\begin{table*}[t]
\centering
\caption{%
Comparison with state-of-the-art alternatives on \textit{GTA5 $\to$ Cityscapes} benchmark. 
The results are averaged over 3 random seeds.
The top performance is highlighted in \textbf{bold} font.
\dag\ indicates that the corresponding framework uses a CNN-based structure.
\ddag\ indicates that the corresponding framework uses a Transformer-based structure.
}
\label{tab:domain_gta}
\vspace{-8pt}
\setlength{\tabcolsep}{2.5pt}
\scalebox{0.9}{
\begin{tabular}{l | ccccccccccccccccccc | c}
\toprule
Method &
\rotatebox{90}{Road} & \rotatebox{90}{S.walk} & \rotatebox{90}{Build.} & \rotatebox{90}{Wall} & \rotatebox{90}{Fence} & \rotatebox{90}{Pole} & \rotatebox{90}{T.light} & \rotatebox{90}{Sign} & \rotatebox{90}{Veget.} & \rotatebox{90}{Terrain} & \rotatebox{90}{Sky} & \rotatebox{90}{Person} & \rotatebox{90}{Rider} & \rotatebox{90}{Car} & \rotatebox{90}{Truck} & \rotatebox{90}{Bus} & \rotatebox{90}{Train} & \rotatebox{90}{M.bike} & \rotatebox{90}{Bike} & mIoU \\
\midrule
source only$^{\dag}$ & 70.2 & 14.6 & 71.3 & 24.1 & 15.3 & 25.5 & 32.1 & 13.5 & 82.9 & 25.1 & 78.0 & 56.2 & 33.3 & 76.3 & 26.6 & 29.8 & 12.3 & 28.5 & 18.0 & 38.6 \\
\midrule
DAFormer$^{\dag}$ & 94.6 & 66.5 & 87.9 & 39.5 & 33.7 & 38.5 & 49.6 & 60.0 &  88.0 & \textbf{46.6} & 88.3 & \textbf{69.6} & 44.4& \textbf{89.0}&46.8&\textbf{56.8}&0.0&17.8&44.3 & 55.9\\
DAFormer (\textit{w/} \method)& \textbf{94.9} & \textbf{68.2} & \textbf{88.8} & \textbf{40.9} & \textbf{37.1} & \textbf{42.6} & \textbf{52.1} & \textbf{62.1} & \textbf{88.3} & 43.3 & \textbf{89.3} & 68.6 & \textbf{44.5}&88.9&\textbf{56.0}&54.6&\textbf{3.8}&\textbf{38.6}&\textbf{58.3}&\textbf{59.0}\\
\midrule
DAFormer$^\ddag$
& 95.7 & 70.2    & \textbf{89.4}     & 53.5 & 48.1  & 49.6 & \textbf{55.8}  & 59.4 & \textbf{89.9}       & 47.9   & \textbf{92.5} & 72.2   & \textbf{44.7}  & \textbf{92.3} & 74.5  & \textbf{78.2} & 65.1   & 55.9      & 61.8 & 68.3 \\
DAFormer (\textit{w/} \method)
& \textbf{96.2}	&\textbf{73.1}&	89.3&	\textbf{53.6}&	\textbf{55.7}&	\textbf{50.9}&	55.7&	\textbf{61.1}&	89.7&	\textbf{52.4}&	92.3&	\textbf{74.7}&	43.5&	91.6&	\textbf{74.6}&	77.4&	\textbf{69.2}&	\textbf{58.9}&	\textbf{62.3}& \textbf{69.6}
 \\ 
\midrule
HRDA$^{\ddag}$& 96.4 & 74.4 & 91.0 & \textbf{61.6} & 51.5 & 57.1 & \textbf{63.9} & 69.3 & 91.3 & 48.4 & 94.2 & \textbf{79.0} & 52.9 & \textbf{93.9} & \textbf{84.1} & \textbf{85.7} & 75.9 & 63.9 & 67.5 & 73.8\\
HRDA (\textit{w/} \method)&\textbf{96.7} &\textbf{76.6} &\textbf{91.5} &61.2 &\textbf{56.9} &\textbf{59.4} &62.2 &\textbf{72.8} &\textbf{91.5} &\textbf{51.2} &\textbf{94.3} &77.5 &\textbf{54.7} &93.5 & 83.2 & 84.7 & \textbf{79.7} &\textbf{68.1} &\textbf{67.6}
&\textbf{74.9}
 \\ 
\bottomrule
\end{tabular}}
%\vspace{-5pt}
\end{table*}

\begin{table*}[t]
\centering
\caption{%
Comparison with state-of-the-art alternatives on \textit{SYNTHIA $\to$ Cityscapes} benchmark. 
The results are averaged over 3 random seeds.
The mIoU and the mIoU* indicate we compute mean IoU over 16 and 13 categories, respectively.
The top performance is highlighted in \textbf{bold} font.
\dag\ indicates that the corresponding framework uses a CNN-based structure.
\ddag\ indicates that the corresponding framework uses a Transformer-based structure.
}
\label{tab:domain_synthia}
\vspace{-8pt}
\setlength{\tabcolsep}{3.5pt}
\scalebox{0.9}{
\begin{tabular}{l | cccccccccccccccc | cc}
\toprule
Method &
\rotatebox{90}{Road} & \rotatebox{90}{S.walk} & \rotatebox{90}{Build.} & \rotatebox{90}{Wall*} & \rotatebox{90}{Fence*} & \rotatebox{90}{Pole*} & \rotatebox{90}{T.light} & \rotatebox{90}{Sign} & \rotatebox{90}{Veget.} & \rotatebox{90}{Sky} & \rotatebox{90}{Person} & \rotatebox{90}{Rider} & \rotatebox{90}{Car} & \rotatebox{90}{Bus} & \rotatebox{90}{M.bike} & \rotatebox{90}{Bike} & mIoU & mIoU* \\
\midrule
% \multicolumn{1}{l}{\textit{}} &
%  \multicolumn{15}{c}{DeepLab-V2~\cite{chen2017deeplab} with ResNet-101~\cite{he2016deep}} \\
% \midrule
source only$^{\dag}$ & 55.6 & 23.8 & 74.6 & 9.2 & 0.2 & 24.4 & 6.1 & 12.1 & 74.8 & 79.0 & 55.3 & 19.1 & 39.6 & 23.3 & 13.7 & 25.0 & 33.5 & 38.6 \\
\midrule
DAFormer$^\dag$  
&62.2 & 24.5 & 85.3 & 23.4 & 2.5 & 38.5 & 47.7 & \textbf{51.1} & 84.0 & 81.8 & 70.5 & \textbf{41.3} & 77.9 & 46.6  & 45.3 & 60.3 & 52.7 & 59.9
\\
DAFormer (\textit{w/} \method)  
&\textbf{70.4}	&\textbf{28.9}&	\textbf{89.2}&	\textbf{25.2}&	\textbf{19.9}&	\textbf{40.2}&	\textbf{55.2}&	50.3&	\textbf{86.9}&	\textbf{84.2}&	\textbf{76.4}&	40.5&	\textbf{79.6}&	\textbf{51.3}&	\textbf{49.2}&	\textbf{61.2}&	\textbf{56.8}& \textbf{63.3}

\\ 
\midrule
% \multicolumn{1}{l}{\textit{Transformer-based}} & \multicolumn{17}{c}{DAFormer~\cite{hoyer2022daformer} with MiT-B5~\cite{xie2021segformer}} \\
% \midrule
%
DAFormer$^\ddag$
& 84.5 & 40.7   & 88.4    & 41.5 & \textbf{6.5}   & 50.0 & 55.0  & 54.6 & 86.0 & 89.8 & 73.2 & \textbf{48.2}  & 87.2 & 53.2 & 53.9 & 61.7 & 60.9 & 67.4\\
DAFormer (\textit{w/} \method) &
\textbf{86.7}&	\textbf{44.9}&	\textbf{89.0}&	\textbf{43.2}&	6.4&	\textbf{52.1}&	\textbf{60.0}&	\textbf{54.9}&	\textbf{88.2}&	\textbf{91.3}&	\textbf{74.9}&	46.1	&\textbf{88.6}&	\textbf{55.6}&	\textbf{55.0}&	\textbf{62.3}&	\textbf{62.5}&\textbf{69.0}\\ 
\midrule
HRDA$^\ddag$
& 85.2 & 47.7 & 88.8 & 49.5 & 4.8 & \textbf{57.2} & 65.7 & 60.9 & 85.3 & 92.9 & \textbf{79.4} & 52.8 & 89.0 & \textbf{64.7} & 63.9 & 64.9 & 65.8 & 72.7 \\
HRDA (\textit{w/} \method)&
\textbf{87.6}&	\textbf{47.9}&	\textbf{90.5}&	\textbf{50.4}&	\textbf{6.9}&	57.1&	64.3&	\textbf{65.3}&	\textbf{86.9}&	\textbf{93.4}&	78.9&	\textbf{54.9}&	\textbf{89.1}&	62.9&	\textbf{65.2}&	\textbf{66.8}&	\textbf{66.8} & \textbf{73.4}
 \\ 
\bottomrule
\end{tabular}}
\vspace{-10pt}
\end{table*}

\noindent\textbf{Datasets.} Unsupervised Domain adaptive (UDA) semantic segmentation aims at transferring the knowledge from a source domain to a target domain.
The source domain is a labeled dataset obtained from the synthetic images and the target domain is an unlabeled real image dataset.
We use two synthetic datasets: GTA5~\cite{richter2016playing_GTA} and SYNTHIA~\cite{ros2016synthia} as source domains respectively and use real images from Cityscapes~\cite{cityscapes} as the target domain.
In other words, We conduct experiments on two dataset settings: \textit{GTA5 $\to$ Cityscapes} and \textit{SYNTHIA $\to$ Cityscapes}.
It is worth mentioning that in \textit{SYNTHIA $\to$ Cityscapes}, $16$ and $13$ of the $19$ classes of Cityscapes are used to calculate mIoU, following the common practice~\cite{hoyer2022daformer}

\noindent\textbf{Implementation details.} We use the recent state-of-the-art frameworks DAFormer~\cite{hoyer2022daformer} and HRDA~\cite{hoyer2022hrda} as the baseline.
In addition, since DAFormer~\cite{hoyer2022daformer} is a pure Transformer-based framework that integrates some effective training strategies, to verify our method, we also try to replace the model structure with ResNet101~\cite{he2016deep} + DeepLabV2~\cite{deeplab}.
This version of DAFormer based on the CNN structure also illustrates the generality of our method.
In all UDA segmenetation experiments, AdamW~\cite{dozat2016incorporating} optimizer is utilized with a learning rate of $6\times 10^{-5}$ for the encoder and $6\times 10^{-4}$ for the decoder.
This optimizer is set to be with a weight decay of $0.01$ along with a linear learning rate warmup with $1.5k$ iterations and linear decay afterward.
During training, for the DAFormer-based methods, per batch input is set to be of two $512\times 512$ random crops.
For HRDA~\cite{hoyer2022hrda}, whose main motivation considers the training image resolution, we adopt the settings consistent with the paper.

\begin{figure}[t]
    \centering
    \includegraphics[width=1.0\linewidth]{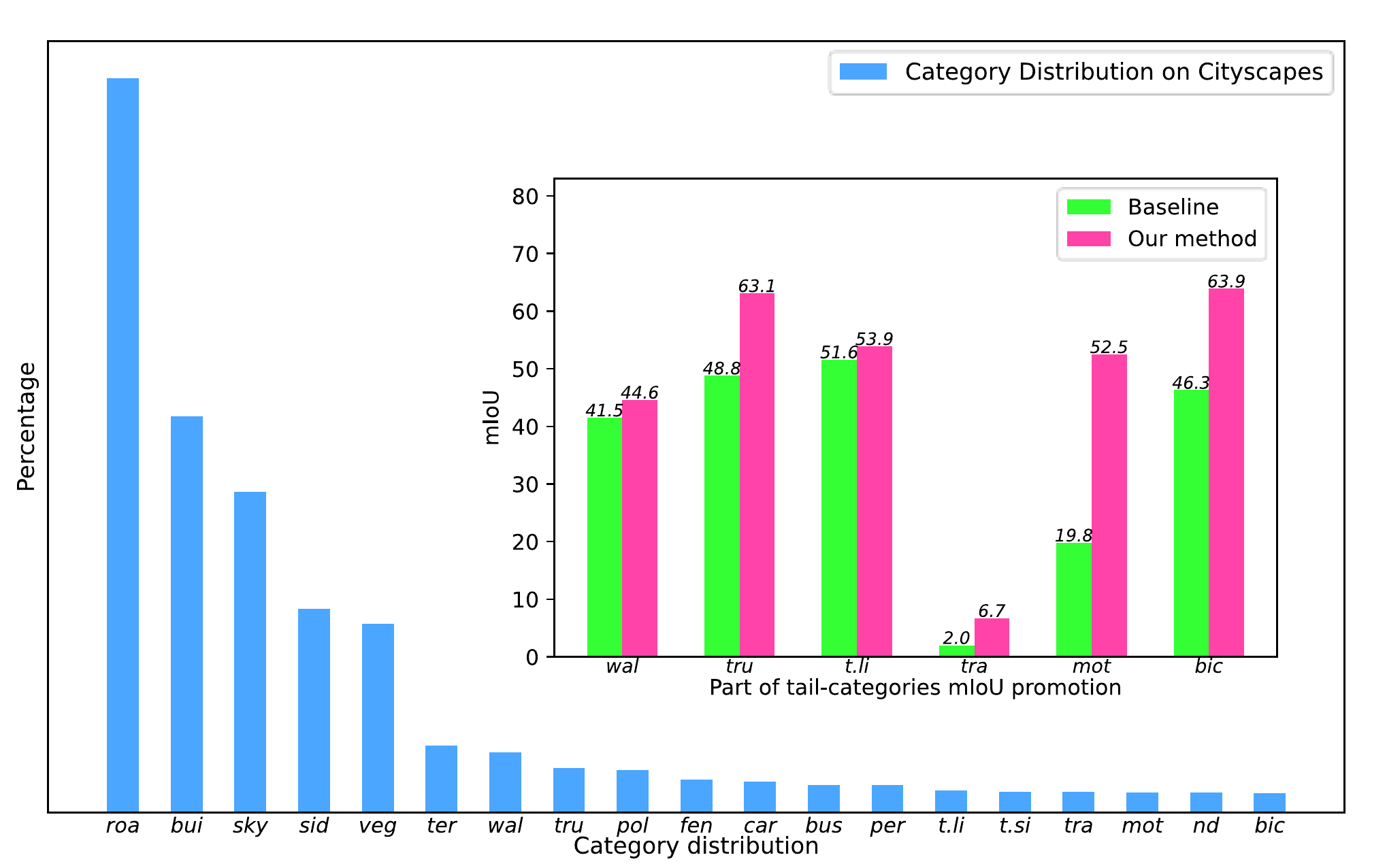}
    \vspace{-15pt}
    \caption{%
        \textbf{Pixel-level category distribution versus performance improvement on multiple tail-categories.} This figure suggests that the performance improvement of our \method comes mainly from the tail category.
    }
    \label{fig:tail}
    \vspace{-18pt}
\end{figure}

\noindent\textbf{Results.} \cref{tab:domain_gta} and \cref{tab:domain_synthia} both suggest that our proposed \method can consistently improve the performance of the UDA segmentation task.
Our \method advances the baseline frameworks DAFormer$^\dag$, DAFormer$^\ddag$, HRDA$^\ddag$ with $+3.1\%$, $+1.3\%$, $+1.1\%$ respectively on \textit{GTA5 $\to$ Cityscapes} benchmark.
\method also advances DAFormer$^\dag$, DAFormer$^\ddag$, HRDA$^\ddag$ with $+4.1\%$, $+1.6\%$, $+1.0\%$ on the mIoU evaluation of 16 categories and  with $+3.4\%$, $+1.6\%$, $+0.7\%$ on the mIoU evaluation of 13 categories respectively on \textit{SYNTHIA $\to$ Cityscapes} benchmark.
From the per-category results in \cref{tab:domain_synthia} and \cref{tab:domain_gta}, we can observe that the improvement of our method for the overall mIoU comes from the improvement of IoU of the tail categories.
We make this conclusion even more obvious by plotting the pixel-level category frequency versus performance improvement on tail categories in figure~\cref{fig:tail}.
Our \method achieves $+3.1\%$, $+14.3\%$, $+2.3\%$, $+4.7\%$, $+32.7\%$ and $+17.6\%$ boosts on ``wall'', ``truck'', ``traffic light'', ``trailer'', ``motorcycle'' and ``bicycle'', respectively, which happen to belong to the tail categories, indicating that our method optimizes the feature space of the tail category which leads to consistent performance improvements.

\subsection{Comparisons with Long-tailed Methods}\label{sec:comparision}
\noindent \textbf{Implementation details.}  
For the fully-supervised setting, we implement Lovász-Softmax~\cite{berman2018lovasz} and Logit-Adjustment~\cite{menon2021long} loss on the ViT-B/16 + UperHead model.
For the semi-supervised setting, we implement \method and Logit-Adjustment~\cite{menon2021long} on the ResNet-50 + PSPNet model for a fair comparison with DARS~\cite{he2021re}.
For the UDA setting, we compare \method with naive resampling the input instances, CBST~\cite{li2022class}, CLAN~\cite{luo2019taking} and Logit-Adjustment~\cite{menon2021long} on \textit{GTA5 $\to$ Cityscapes}. 
\noindent \textbf{Results.}
% \todo{We will add more complete comparisons in the revision}.
We demonstrate more comparisons in \cref{tab:more_comparision}.
\method achieves the mIoU of $77.7\%$ on fully-supervised setting, $73.2\%$ on semi-supervised setting and $59.0\%$ on the domain adaptive setting.
\method also achieves the mIoU of $66.2\%$ on fully-supervised setting, $59.3\%$ on semi-supervised setting, and $45.7\%$ on the domain adaptive setting for the tail-categories.
The overall better performance suggests \method boosts the tail categories and outperforms other alternatives on different settings and benchmarks, which demonstrates its versatility and effectiveness.

\subsection{Ablation Studies}\label{sec:ablation}

\noindent \textbf{Exploration on the form of variations.} In \cref{tab:abla_1}, we explore the influence of different perturbation forms on the experimental results.
Results are obtained from the DAFormer$\dag$ framework on \textit{GTA5 $\to$ Cityscapes setting}.
``None-Variation'' denotes the pure baseline. 
``Gaussian'' variation parameters are illustrated in \cref{sec:method},
For the ``Uniform'' term, We sample uniformly in the interval $\left[0, 1\right]$, which brings in the baseline with a boost of $+2.3\%$.
For the ``Beta'' variation, we set the $\alpha=0.5$ and $\beta=0.5$, which advances the baseline by $+2.0\%$.
For the ``Exponential'' variation, we set the $\lambda=1$, which surpasses the baseline by $+2.6\%$ 
In order to ensure that the size of the perturbation is within a reasonable range, we clip all perturbations to the $\left[0, 1\right]$ interval.
We empirically find that the ``Gaussian'' variation term outperforms all the other alternatives with a mIoU of $59.0\%$, while all the other variations advance the baseline.
These suggest that various variations can improve the performance of the task, and the key to the improvement lies in the coefficients related to the category frequency rather than in the form of variation.
This is more in line with our intuition.
If the parameters of other variations are finetuned, better results may be obtained.

\noindent \textbf{Exploration on the variance $\sigma$.} As the only hyper-parameter needs to be carefully tuned, we ablate $\sigma$ for potential generalized usage extended to other tasks. 
\cref{tab:ablation_k} gives experimental results on the influence of different $\sigma$ with DAFormer $\dag$ under two different adaptation settings: \textit{GTA5 $\to$ Cityscapes} and \textit{SYNTHIA $\to$ Cityscapes}. 
\method advances the baseline most by $+3.1\%$ when $\sigma=6$ for \textit{GTA5 $\to$ Cityscapes} and  by $+4.1\%$ when $\sigma=4$ for \textit{SYNTHIA $\to$ Cityscapes}.
Although the different choices of $\sigma$ affect the final performance, these gaps are quite trivial(the discrepancy between the maximum and minimum mIoU is within $+1\%$), which demonstrates that our \method is robust to hyper-parameter choices to some extent and indicates its good scalability.

\begin{table}[t]
    \setlength{\tabcolsep}{5pt}
    \centering
    \caption{%
            More comparisons with other long-tailed baselines on different semantic segmentation tasks. 
            ``RS'' and ``RW'' denotes the naive resample and reweight trick respectively.
            $^{\ast}$ means the results come from our implementation.
            $^{\circ}$ means the results come from the original papers.
    }
    \label{tab:more_comparision}
    \vspace{-5pt}
    \scalebox{0.8}{
    \begin{tabular}{l|llllll}
        \toprule
    Supervision&\multicolumn{3}{c}{Fully} & \multicolumn{3}{|c}{Semi}  \\
       \midrule
Method & LA$^{\ast}$ &Lovász$^{\ast}$ & \method &\multicolumn{1}{|l}{LA$^{\ast}$}&DARS$^{\circ}$ & \method  \\
    \midrule
    mIoU&75.9 &76.6 & \textbf{77.7} & \multicolumn{1}{|l}{69.3} &72.8 &\textbf{73.2} \\
 mIoU (tail)&62.4  & 63.9 & \textbf{66.2} & \multicolumn{1}{|l}{55.7} & 58.4&\textbf{59.3}  \\
   \midrule
   Supervision&\multicolumn{6}{c}{Domain Adaptive}\\
   \midrule
   Method  &RS &RW &CLAN$^{\circ}$ & CBST$^{\circ}$ & LA$^{\ast}$ & \method \\
   \midrule
    mIoU& 56.2& 56.4&43.2 &45.9 &56.5 & \textbf{59.0} \\
     mIoU (tail)&40.5  &40.8  & 25.9 &28.5 &41.9 &\textbf{45.7} \\
%    mIoU(tail)& & 28.5 & 41.9 & \textbf{45.7} \\
        % U$^2$PL & 
        % \textbf{74.90} \scriptsize{(\textcolor{blue}{$+$0.45})} & \textbf{76.48} \scriptsize{(\textcolor{blue}{$+$0.93})} & \textbf{78.51} \scriptsize{(\textcolor{blue}{$+$1.03})} & \textbf{79.12} \scriptsize{(\textcolor{blue}{$+$0.11})} \\
        \bottomrule
    \end{tabular}
    }
    \vspace{-5pt}
\end{table}

\begin{table}[t]
\centering
\caption{%
\textbf{Ablation study on various types of variations.} ``None-Variation'' denotes the DAFormer$\dag$ baseline.
The green arrows indicate the relative improvement in performance.
}
\label{tab:abla_1}
\vspace{-5pt}
\setlength{\tabcolsep}{22pt}
\scalebox{1}{
\begin{tabular}{l | c  c}
\toprule
{Variation}&\multicolumn{2}{c}{mIoU} \\
\midrule
None-Variation & 55.9 &\\
Gaussian& \textbf{59.0}&\textbf{\up{3.1}} \\
Uniform & 58.2&\textbf{\up{2.3}}\\
Beta & 57.9&\textbf{\up{2.0}}\\
Exponential & 58.5&\textbf{\up{2.6}} \\

\bottomrule
\end{tabular}}
\vspace{-5pt}
\end{table}

\begin{table}[t]
\centering
\caption{%
\textbf{Ablation study on the variance $\sigma$ in \cref{eq:balancing},}
which determines the overall magnitude of the variation.
}
\setlength{\tabcolsep}{10pt}
\label{tab:ablation_k}
\vspace{-5pt}
\setlength{\tabcolsep}{9pt}
\begin{tabular}{cccccc}
\toprule
Baseline    &     3& 4 & 5 & 6 & 7   \\
\midrule
\multicolumn{6}{c}{\textit{GTA5 $\to$ Cityscapes}}
\\
55.9  & 58.0 &    58.8   &        58.2   &         \textbf{59.0}   &  58.7          \\
\midrule
\multicolumn{6}{c}{\textit{SYNTHIA $\to$ Cityscapes}} \\
52.7  &56.5  &   \textbf{56.8}    &   55.9        & 56.3           &   56.1         \\
\bottomrule
\end{tabular}
\vspace{-5pt}
\end{table}

\begin{table}[t]
\centering
\caption{%
\textbf{Ablation study on the components of \method.} 
We ablate two components in \cref{eq:balancing}.
``\textit{w/o} variation'' indicates removing the $|\delta(\sigma)|$ term.
``\textit{w/o} variation'' indicates removing the pixel-level category frequency term.
}
\setlength{\tabcolsep}{10pt}
\label{tab:ablation_components}
\vspace{-5pt}
\begin{tabular}{cccc}
\toprule
 Baseline & \textit{w/o} variation & \textit{w/o} balance& \method   \\
\midrule
\multicolumn{4}{c}{\textit{GTA5 $\to$ Cityscapes}} \\
 55.9 &   56.5      &    56.8 &    \textbf{59.0}          \\
\midrule
\multicolumn{4}{c}{\textit{SYNTHIA $\to$ Cityscapes}} \\
 52.7 &   53.9    &  54.5         &     \textbf{56.8}    \\
  
\bottomrule
\end{tabular}
\vspace{-10pt}
\end{table}

\begin{figure*}[t]
    \centering
    \includegraphics[width=1\textwidth]{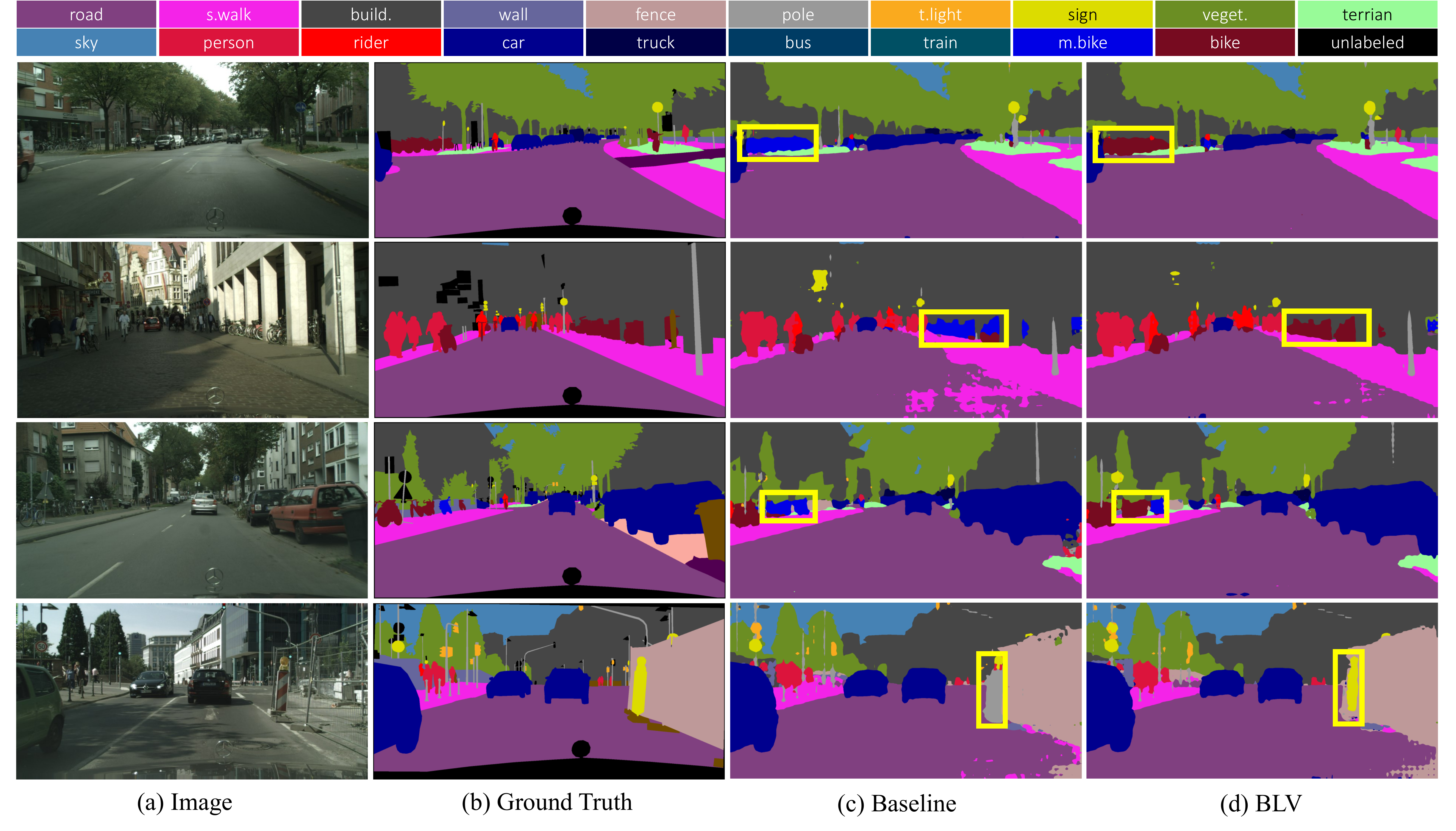}
     \vspace{-18pt}
    \caption{%
    \textbf{Qualitative results on Cityscapes val set}.
    Baseline and BLV are trained on \textit{GTA5$\to$ Cityscapes} benchmark of unsupervised domain adaptive semantic segmentation task.
    (a) Input images. 
    (b) Ground truth annotations for the corresponding images.
    (c) Result of the DAFormer baseline.
    (d) Result of our method (DAFormer + BLV).
    Yellow rectangles highlight the promotion of segmentation results by our method on tail categories.
    }
    \label{fig:vis}
    \vspace{-10pt}
\end{figure*}

\noindent \textbf{Exploration on components of \method.} Corresponding results are demonstrated in \cref{tab:ablation_components}.
``$\textit{w/o}$ variation'' denotes our \method without adding the variation in \cref{sec:method} rather a constant value adjustment for each category.
``$\textit{w/o}$ balance'' denotes our \method without adding categorical balance coefficient in \cref{sec:method} rather a fixed-scale adjustment for each category.
\cref{tab:ablation_components} demonstrates that either the ``$\textit{w/o}$ variation'' or ``$\textit{w/o}$ balance'' can boost the baseline non-trivially on both the \textit{GTA5 $\to$ Cityscapes} by $+0.6\%$, $+0.9\%$ and the \textit{SYNTHIA $\to$ Cityscapes} settings by $+1.2\%$, $+1.8\%$, although not as significant as our \method.
This suggests two conclusions: 
1) Adding category-agnostic consistent variation to logits can indeed optimize the representation space to a certain extent, but it cannot completely solve the adverse effects of long-tailed data.
2) Adding static category-related adjustments can also alleviate this problem, but this cannot enrich training instances thus leading to potential overfitting problems while the variation term of \method can be used to avoid this problem efficiently.
This ablation experiment demonstrates the necessity of all components of our proposed \method.

\subsection{Visualization}
\cref{fig:vis} shows the result of our method on the Cityscapes val set.
With this visualization, we prove that overlaying our method to the baseline is effective in alleviating category confusion, so our method achieves better performance.
More details can be demonstrated by the yellow rectangle highlighting part in \cref{fig:vis}c and \cref{fig:vis}d. (\textit{i.e.} the pixel misclassified in the baseline are corrected by balancing logit variation.)

%% file: sections/5.conclusion.tex
\section{Conclusion}\label{sec:conclusion}
In this paper, we propose the \method, a simple yet effective plug-in design for various kinds of long-tail semantic segmentation tasks. 
We introduce category scale-related variation during the model training stage.
This variation is inversely proportional to the frequency of occurrences of instances, which effectively closes the gap between the feature area of different categories.
Extensive experiments on fully supervised, semi-supervised, and unsupervised domain adaptive semantic segmentation tasks suggest our method can boost performance.
Compared with other methods for alleviating the class-imbalance issues, our \method is better and more concise and general.
Furthermore, sufficient ablation experiments as well as intuitive visualization results prove the necessity of individual components and the effectiveness of our method.
%
%\method has the potential to be extended to other tasks and become a general paradigm.

\noindent\textbf{Discussion.} One necessary premise of \method is that the category frequencies need to be known.
It is unlikely to be satisfied in some tasks like unsupervised semantic segmentation and  domain generalized semantic segmentation.
%
% Besides, this method may involve privacy issues of training data.

% Although this premise can be approximated by the source domain data category distribution in UDA tasks,
%

%% file: sections/6.ref.tex
{\small
\bibliographystyle{ieee_fullname}
\bibliography{ref}
}

%% file: sections/7.supple.tex
% \noindent \textcolor{red}{$\left[Typo\ correction\right]$}
% \noindent  \textcolor{red}{$\bullet$ Eq. (3) in the main paper should be corrected into 
% \begin{equation}\label{eq:sup_balancing}
% \hat{z}^{i}_{k} = z^{i}_{k} + \frac{c_{k}}{\max_{i=0}^{C-1}(c_{i})}|\delta(\sigma)|,\quad c_{k} =\log \frac{\sum_{j=0}^{C-1}q_{j}}{q_{k}}
% \end{equation}}
% \noindent  \textcolor{red}{$\bullet$ The ``Exploration on the variance $k$" in the ``Ablation Studies" section should be corrected into ``Exploration on the variance $\sigma$"} 

\definecolor{myblue}{RGB}{0,0,255}

\section{Overview}
In this supplementary material, we first provide more details for reproducibility in \cref{sec-sup:reprodu}.
We further explore a potential improvement of \method and corresponding preliminary results in \cref{sec-sup:more-expro}.
Then we demonstrate our \method with more UDA methods on both the \textit{GTA5 $\to$ Cityscapes} and \textit{SYNTHIA $\to$ Cityscapes} settings in \cref{sec-sup:comp-uda}.
Intuitive feature space visualization is demonstrated by t-SNE method in \cref{sec-sup:tsne-visual}.
Pseudo-code for direct understanding of \method is provided in \cref{sec-sup:pseudo-code}.
Information about computational overhead and distribution estimation is exhibited in \cref{sec-sup:training-time} and \cref{sec-sup:estimation-labeled}.

\section{More Details for Reproducibility}\label{sec-sup:reprodu}
\noindent \textbf{Details for parameters.}
As we mentioned in the paper, the only parameter for \method is the $\sigma$ in Eq. (\textcolor{red}{3}).

We set $\sigma=4$ for unsupervised domain adaptive semantic segmentation task under the \textit{SYNTHIA $\to$ Cityscapes} setting.
For all the other tasks, we set $\sigma=6$ consistently.
Besides, the $\delta(\sigma)$ term is clamped into $\left[0, 1\right]$ to avoid particularly large values that makes training unstable.

\noindent \textbf{Details for data augmentation.}
We follow DACS~\cite{tranheden2021dacs}, using color jitter, Gaussian blur, and ClassMix~\cite{classmix} as the augmentation selections.

\section{More Exploration of Variation} \label{sec-sup:more-expro}

\begin{figure}[ht]
    \centering
    \includegraphics[width=1.0\linewidth]{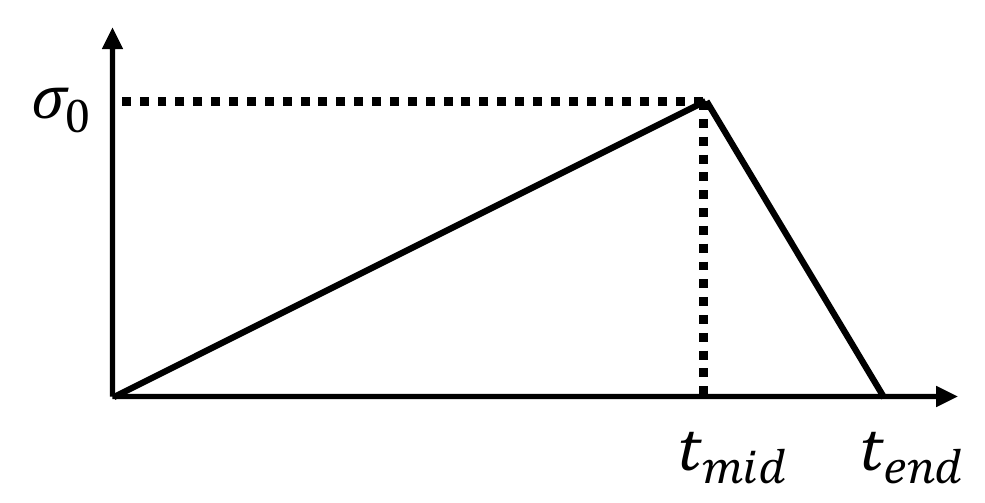}
    \vspace{-15pt}
    \caption{
        \textbf{$\sigma$ that changes with training iterations.}
        $t_{end}$ is the total iterations.
        $t_{mid}$ is the turning point of $\sigma$ with a corresponding maximum value $\sigma_{0}$.
    }
    \label{fig:more_variation}
    \vspace{-10pt}
\end{figure}

We explore the improvement over \method.
We set the $\sigma$ in Eq. (\textcolor{red}{3}) as a temporal variable: $\sigma(t)$, where $t$ denotes current iteration, $t_{mid}$ and $\sigma_{0}$ are hyper-parameters with preset values.
\cref{fig:more_variation} depicts how $\sigma$ changes with iterations.

The main idea is to let the perturbation increase gradually before $t_{mid}$ to obtain an effective variation. 
After $t_{mid}$, we should let the variation decrease so that the model convergence is not affected.
This exploration is easy to implement. 
We conducted the experiment under \textit{GTA5 $\to$ Cityscapes} benchmark.
$t_{end}=40k$, $t_{mid}=30k$ and $\sigma_{0}=6$.

\begin{table}[!ht]
\centering
\caption{
\textbf{Exploration on temporal variation of \method.} ``+tv" denotes our proposed ``temporal variable".
}
\setlength{\tabcolsep}{16pt}
\label{tab:sup_explor}
\vspace{-8pt}
\begin{tabular}{ccc}
\toprule
 Baseline &  \method &  \textbf{\method+tv}  \\
\midrule
 68.3 &   69.6\textbf{\up{1.3}}        &    \textbf{70.0\up{1.7}}     \\
\bottomrule
\end{tabular}
\vspace{-10pt}
\end{table}

The result is demonstrated in ~\cref{tab:sup_explor}.
The baseline is DAFormer$\ddag$ model.
This table suggests that this ``temporal variable" does improve the original \method.
The overall result indicates that there is an opportunity to further improve our approach.

\section{More Comparisons on UDA Benchmark}  \label{sec-sup:comp-uda}
%
% Most studies use CNN as the backbone.
% %
% In this section, we also compare category performance of our method with other state-of-the-art CNN-based methods.
% %
% As shown in \cref{tab:sup-udaperformance}, our \method with ResNet-101 achieves competitive performance among existing methods.
% %
% Note that, we report the performances of ProDA \cite{zhang2021prototypical} and CPSL \cite{li2022class} in \cref{tab:sup-udaperformance} \textit{without} knowledge distillation (which uses self-supervised trained models) for a fair comparison.
% %
% On the SYNTHIA $\to$ Cityscapes benchmark, we set $\mu'=0.9999$ for our \method.
% As shown in \cref{tab:supsynthiaperformance}, our method also demonstrates competitive performance, which is slightly lower than CPSL, a class-balanced training approach that is orthogonal to our work.
%
%%%%%%%%%%%%%%%%%%%%%%%%%%%%%%%%%%%%%%%%%%%%%%%%%%%%%%%%%%%%%%%%%
\begin{table*}[t]
\centering
\caption{
Comparison with state-of-the-art alternatives on \textit{GTA5 $\to$ Cityscapes} benchmark with ResNet-101~\cite{he2016deep} and DeepLab-V2~\cite{deeplab}.
The results are averaged over 3 random seeds.
The top performance is highlighted in \textbf{bold} font and the second score is \textit{underlined}.}
\vspace{10pt}
\label{tab:supgta5performance}
\setlength{\tabcolsep}{2.3pt}
\scalebox{0.95}{
\begin{tabular}{l | ccccccccccccccccccc | cc}
\toprule
Method &
\rotatebox{90}{Road} & \rotatebox{90}{S.walk} & \rotatebox{90}{Build.} & \rotatebox{90}{Wall*} & \rotatebox{90}{Fence*} & \rotatebox{90}{Pole*} & \rotatebox{90}{T.light} & \rotatebox{90}{Sign} & \rotatebox{90}{Veget.} & \rotatebox{90}{Terrain} & \rotatebox{90}{Sky} & \rotatebox{90}{Person} & \rotatebox{90}{Rider} & \rotatebox{90}{Car} & \rotatebox{90}{Truck} & \rotatebox{90}{Bus} & \rotatebox{90}{Train} & \rotatebox{90}{M.bike} & \rotatebox{90}{Bike} & mIoU  \\
\midrule
source only & 70.2 & 14.6 & 71.3 & 24.1 & 15.3 & 25.5 & 32.1 & 13.5 & 82.9 & 25.1 & 78.0 & 56.2 & 33.3 & 76.3 & 26.6 & 29.8 & 12.3 & 28.5 & 18.0 & 38.6 \\
\midrule
AdaptSeg~\cite{tsai2018learning} & 86.5 & 36.0 & 79.9 & 23.4 & 23.3 & 23.9 & 35.2 & 14.8 & 83.4 & 33.3 & 75.6 & 58.5 & 27.6 & 73.7 & 32.5 & 35.4 & 3.9 & 30.1 & 28.1 & 41.4 \\
CyCADA~\cite{hoffman2018cycada} & 86.7 & 35.6 & 80.1 & 19.8 & 17.5 & 38.0 & 39.9 & 41.5 & 82.7 & 27.9 & 73.6 & 64.9 & 19.0 & 65.0 & 12.0 & 28.6 & 4.5 & 31.1 & 42.0 & 42.7 \\
ADVENT~\cite{vu2019advent} & 89.4 & 33.1 & 81.0 & 26.6 & 26.8 & 27.2 & 33.5 & 24.7 & 83.9 & 36.7 & 78.8 & 58.7 & 30.5 & 84.8 & 38.5 & 44.5 & 1.7 & 31.6 & 32.4 & 45.5 \\
CBST~\cite{zou2018unsupervised} & 91.8 & 53.5 & 80.5 & 32.7 & 21.0 & 34.0 & 28.9 & 20.4 & 83.9 & 34.2 & 80.9 & 53.1 & 24.0 & 82.7 & 30.3 & 35.9 & 16.0 & 25.9 & 42.8 & 45.9 \\
PCLA~\cite{kang2020pixel} & 84.0 & 30.4 & 82.4 & 35.3 & 24.8 & 32.2 & 36.8 & 24.5 & 85.5 & 37.2 & 78.6 & 66.9 & 32.8 & 85.5 & 40.4 & 48.0 & 8.8 & 29.8 & 41.8 & 47.7 \\
FADA~\cite{wang2020classes} & 92.5 & 47.5 & 85.1 & 37.6 & 32.8 & 33.4 & 33.8 & 18.4 & 85.3 & 37.7 & 83.5 & 63.2 & \underline{39.7} & 87.5 & 32.9 & 47.8 & 1.6 & 34.9 & 39.5 & 49.2 \\
MCS~\cite{chung2022maximizing} & 92.6 & 54.0 & 85.4 & 35.0 & 26.0 & 32.4 & 41.2 & 29.7 & 85.1 & 40.9 & 85.4 & 62.6 & 34.7 & 85.7 & 35.6 & 50.8 & 2.4 & 31.0 & 34.0 & 49.7 \\
CAG~\cite{zhang2019category} & 90.4 & 51.6 & 83.8 & 34.2 & 27.8 & 38.4 & 25.3 & 48.4 & 85.4 & 38.2 & 78.1 & 58.6 & 34.6 & 84.7 & 21.9 & 42.7 & \textbf{41.1} & 29.3 & 37.2 & 50.2 \\
FDA~\cite{yang2020fda} & 92.5 & 53.3 & 82.4 & 26.5 & 27.6 & 36.4 & 40.6 & 38.9 & 82.3 & 39.8 & 78.0 & 62.6 & 34.4 & 84.9 & 34.1 & 53.1 & 16.9 & 27.7 & 46.4 & 50.5 \\
PIT~\cite{lv2020cross} & 87.5 & 43.4 & 78.8 & 31.2 & 30.2 & 36.3 & 39.3 & 42.0 & 79.2 & 37.1 & 79.3 & 65.4 & 37.5 & 83.2 & \underline{46.0} & 45.6 & \underline{25.7} & 23.5 & 49.9 & 50.6 \\
IAST~\cite{mei2020instance} & \underline{93.8} & 57.8 & 85.1 & 39.5 & 26.7 & 26.2 & 43.1 & 34.7 & 84.9 & 32.9 & 88.0 & 62.6 & 29.0 & 87.3 & 39.2 & 49.6 & 23.2 & 34.7 & 39.6 & 51.5 \\
DACS~\cite{tranheden2021dacs} & 89.9 & 39.7 & \underline{87.9} & 30.7 & 39.5 & 38.5 & 46.4 & \underline{52.8} & \underline{88.0} & \textbf{44.0} & \underline{88.8} & 67.2 & 35.8 & 84.5 & 45.7 & 50.2 & 0.0 & 27.3 & 34.0 & 52.1 \\
RCCR~\cite{zhou2021domain} & 93.7 & \underline{60.4} & 86.5 & 41.1 & 32.0 & 37.3 & 38.7 & 38.6 & 87.2 & 43.0 & 85.5 & 65.4 & 35.1 & \underline{88.3} & 41.8 & 51.6 & 0.0 & 38.0 & 52.1 & 53.5 \\  
ProDA~\cite{zhang2021prototypical} & 91.5 & 52.4 & 82.9 & \underline{42.0} & \underline{35.7} & 40.0 & 44.4 & 43.3 & 87.0 & \underline{43.8} & 79.5 & 66.5 & 31.4 & 86.7 & 41.1 & 52.5 & 0.0 & \underline{45.4} & \underline{53.8} & 53.7 \\  
CPSL~\cite{li2022class} & 91.7 & 52.9 & 83.6 & \textbf{43.0} & 32.3 & \textbf{43.7} & \underline{51.3} & 42.8 & 85.4 & 37.6 & 81.1 & \textbf{69.5} & 30.0 & 88.1 & 44.1 & \textbf{59.9} & 24.9 & \textbf{47.2} & 48.4 & \underline{55.7} \\
\midrule
\method (ours) & \textbf{94.9} & \textbf{68.2} & \textbf{88.8} & 40.9 & \textbf{37.1} & \underline{42.6} & \textbf{52.1} & \textbf{62.1} & \textbf{88.3} & 43.3 & \textbf{89.3} & \underline{68.6} & \textbf{44.5} & \textbf{88.9} & \textbf{56.0} & \underline{54.6} & 3.8 & 38.6 & \textbf{58.3} & \textbf{59.0} \\
\bottomrule
\end{tabular}}
\vspace{-5pt}
\end{table*}

\begin{table*}[t]
\centering
\caption{
Comparison with state-of-the-art alternatives on \textit{SYNTHIA $\to$ Cityscapes} benchmark with ResNet-101~\cite{he2016deep} and DeepLab-V2~\cite{deeplab}. 
The results are averaged over 3 random seeds.
The mIoU and the mIoU* indicate we compute mean IoU over 16 and 13 categories, respectively.
The top performance is highlighted in \textbf{bold} font and the second score is \textit{underlined}.}
\vspace{10pt}
\label{tab:supsynthiaperformance}
\setlength{\tabcolsep}{2.3pt}
\scalebox{1}{
\begin{tabular}{l | cccccccccccccccc | cc}
\toprule
Method &
\rotatebox{90}{Road} & \rotatebox{90}{S.walk} & \rotatebox{90}{Build.} & \rotatebox{90}{Wall*} & \rotatebox{90}{Fence*} & \rotatebox{90}{Pole*} & \rotatebox{90}{T.light} & \rotatebox{90}{Sign} & \rotatebox{90}{Veget.} & \rotatebox{90}{Sky} & \rotatebox{90}{Person} & \rotatebox{90}{Rider} & \rotatebox{90}{Car} & \rotatebox{90}{Bus} & \rotatebox{90}{M.bike} & \rotatebox{90}{Bike} & mIoU & mIoU* \\
\midrule
source only$^\dag$ & 55.6 & 23.8 & 74.6 & 9.2 & 0.2 & 24.4 & 6.1 & 12.1 & 74.8 & 79.0 & 55.3 & 19.1 & 39.6 & 23.3 & 13.7 & 25.0 & 33.5 & 38.6 \\
\midrule
AdaptSeg~\cite{tsai2018learning} & 79.2 & 37.2 & 78.8 & - & - & - & 9.9 & 10.5 & 78.2 & 80.5 & 53.5 & 19.6 & 67.0 & 29.5 & 21.6 & 31.3 & - & 45.9 \\
ADVENT~\cite{vu2019advent} & 85.6 & 42.2 & 79.7 & 8.7 & 0.4 & 25.9 & 5.4 & 8.1 & 80.4 & 84.1 & 57.9 & 23.8 & 73.3 & 36.4 & 14.2 & 33.0 & 41.2 & 48.0 \\
CBST~\cite{zou2018unsupervised} & 68.0 & 29.9 & 76.3 & 10.8 & 1.4 & 33.9 & 22.8 & 29.5 & 77.6 & 78.3 & 60.6 & 28.3 & 81.6 & 23.5 & 18.8 & 39.8 & 42.6 & 48.9 \\
CAG~\cite{zhang2019category} & 84.7 & 40.8 & 81.7 & 7.8 & 0.0 & 35.1 & 13.3 & 22.7 & 84.5 & 77.6 & 64.2 & 27.8 & 80.9 & 19.7 & 22.7 & 48.3 & 44.5 & 51.5\\
PIT~\cite{lv2020cross} & 83.1 & 27.6 & 81.5 & 8.9 & 0.3 & 21.8 & 26.4 & 33.8 & 76.4 & 78.8 & 64.2 & 27.6 & 79.6 & 31.2 & 31.0 & 31.3 & 44.0 & 51.8 \\
FDA~\cite{yang2020fda} & 79.3 & 35.0 & 73.2 & - & - & - & 19.9 & 24.0 & 61.7 & 82.6 & 61.4 & 31.1 & 83.9 & 40.8 & 38.4 & 51.1 & - & 52.5 \\
FADA~\cite{wang2020classes} & 84.5 & 40.1 & 83.1 & 4.8 & 0.0 & 34.3 & 20.1 & 27.2 & 84.8 & 84.0 & 53.5 & 22.6 & 85.4 & 43.7 & 26.8 & 27.8 & 45.2 & 52.5 \\
MCS~\cite{chung2022maximizing} & \underline{88.3} & \textbf{47.3} & 80.1 & - & - & - & 21.6 & 20.2 & 79.6 & 82.1 & 59.0 & 28.2 & 82.0 & 39.2 & 17.3 & 46.7 & - & 53.2 \\
PyCDA~\cite{lian2019constructing} & 75.5 & 30.9 & 83.3 & 20.8 & 0.7 & 32.7 & 27.3 & 33.5 & 84.7 & 85.0 & 64.1 & 25.4 & 85.0 & 45.2 & 21.2 & 32.0 & 46.7 & 53.3 \\
PLCA~\cite{kang2020pixel} & 82.6 & 29.0 & 81.0 & 11.2 & 0.2 & 33.6 & 24.9 & 18.3 & 82.8 & 82.3 & 62.1 & 26.5 & 85.6 & 48.9 & 26.8 & 52.2 & 46.8 & 54.0 \\
DACS~\cite{tranheden2021dacs} & 80.6 & 25.1 & 81.9 & 21.5 & 2.9 & 37.2 & 22.7 & 24.0 & 83.7 & \textbf{90.8} & 67.6 & \underline{38.3} & 82.9 & 38.9 & 28.5 & 47.6 & 48.3 & 54.8 \\
RCCR~\cite{zhou2021domain} & 79.4 & 45.3 & 83.3 & - & - & - & 24.7 & 29.6 & 68.9 & 87.5 & 63.1 & 33.8 & 87.0 & 51.0 & 32.1 & 52.1 & - & 56.8 \\
IAST~\cite{mei2020instance} & 81.9 & 41.5 & 83.3 & 17.7 & \underline{4.6} & 32.3 & 30.9 & 28.8 & 83.4 & 85.0 & 65.5 & 30.8 & 86.5 & 38.2 & 33.1 & 52.7 & 49.8 & 57.0 \\
ProDA~\cite{zhang2021prototypical} & 87.1 & 44.0 & 83.2 & \textbf{26.9} & 0.7 & 42.0 & 45.8 & \underline{34.2} & 86.7 & 81.3 & 68.4 & 22.1 & \underline{87.7} & 50.0 & 31.4 & 38.6 & 51.9 & 58.5 \\
SAC~\cite{araslanov2021self} & \textbf{89.3} & \underline{47.2} & \underline{85.5} & \underline{26.5} & 1.3 & \textbf{43.0} & 45.5 & 32.0 & \textbf{87.1} & \underline{89.3} & 63.6 & 25.4 & 86.9 & 35.6 & 30.4 & 53.0 & 52.6 & 59.3 \\
CPSL~\cite{li2022class} & 87.3 & 44.4 & 83.8 & 25.0 & 0.4 & \underline{42.9} & \underline{47.5} & 32.4 & 86.5 & 83.3 & \underline{69.6} & 29.1 & \textbf{89.4} & \textbf{52.1} & \underline{42.6} & \underline{54.1} & \underline{54.4} & \underline{61.7} \\ 
\midrule
\method (ours) &
70.4	&28.9&	\textbf{89.2}&	25.2&	\textbf{19.9}&	40.2&	\textbf{55.2}&	\textbf{50.3}&	\underline{86.9}&	84.2&	\textbf{76.4}&	\textbf{40.5}&	79.6&	\underline{51.3}&	\textbf{49.2}&	\textbf{61.2}&	\textbf{56.8}& \textbf{63.3}\\ 
\bottomrule
\end{tabular}}
\vspace{-5pt}
\end{table*}

We add more comparisons of \method with previous UDA methods for GTA5 $\to$ Cityscapes in
\cref{tab:supgta5performance} and for SYNTHIA $\to$ Cityscapes in \cref{tab:supsynthiaperformance}.

We include following methods for comparision: AdaptSeg~\cite{tsai2018learning}, CyCADA~\cite{hoffman2018cycada},  ADVENT~\cite{vu2019advent}, FADA~\cite{wang2020classes}, CBST~\cite{zou2018unsupervised}, IAST~\cite{mei2020instance}, CAG~\cite{zhang2019category}, ProDA~\cite{zhang2021prototypical}, SAC~\cite{araslanov2021self}, CPSL~\cite{li2022class},
PLCA~\cite{kang2020pixel}, RCCR~\cite{zhou2021domain}, and MCS~\cite{chung2022maximizing}.
All methods in \cref{tab:supgta5performance} and \cref{tab:supsynthiaperformance} are based on ResNet-101~\cite{he2016deep} + DeepLab V2~\cite{deeplab}.

\method surpasses other alternatives by a large margin, achieving mIoU of $59.0\%$ on \textit{GTA5 $\to$ Cityscapes}, and $56.8\%$ over 16 classes and $63.3\%$ over 13 classes on \textit{SYNTHIA $\to$ Cityscapes}, respectively.

\section{Visualization on Feature Space} \label{sec-sup:tsne-visual}
\begin{figure}[ht]
    \centering
    \includegraphics[width=1.0\linewidth]{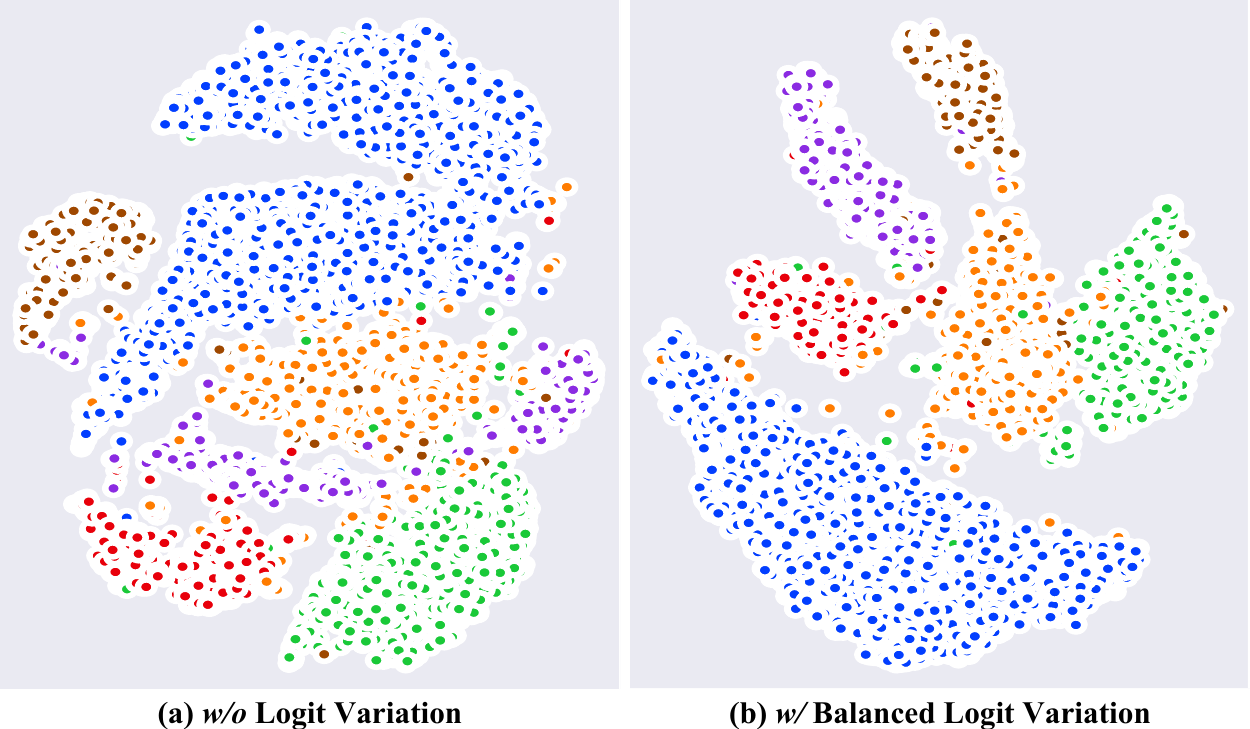}
    \vspace{-15pt}
    \caption{
        \textbf{t-SNE visualization from the logit feature space.}
        (a) Without logit variation, the spacing between instances of different categories is small, resulting in easy misclassification.
        (b) With balanced logit variation, instances are easier to distinguish.
    }
    \label{fig:tsne-visual}
    \vspace{-10pt}
\end{figure}

We use t-SNE~\cite{van2008visualizing} to visualize the logit feature space in ~\cref{fig:tsne-visual}.
In terms of the degree of confusion in the feature space, \method improves the baseline and proves its superiority.

\section{Pseudo-code} \label{sec-sup:pseudo-code}
To make \method easy to understand, we provide pseudo-code in a Pytorch-like style in \cref{alg:code}.
%
% algorithm
\begin{algorithm}[ht]
    \caption{Pseudo-code of \method in a PyTorch-like style.}
    \label{alg:code}
    \definecolor{codeblue}{rgb}{0.25,0.5,0.5}
    \lstset{
      backgroundcolor=\color{white},
      basicstyle=\fontsize{7.2pt}{7.2pt}\ttfamily\selectfont,
      columns=fullflexible,
      breaklines=true,
      captionpos=b,
      commentstyle=\fontsize{7.2pt}{7.2pt}\color{codeblue},
      keywordstyle=\fontsize{7.2pt}{7.2pt},
    %  frame=tb,
    }
    \begin{lstlisting}[language=python]
# frequency_list: a list containing the frequency of pixels of each category.
# pred: model output logits
# target: ground-truch label
# sigma: hyper-paramter

def BLV_Loss(pred, target, sigma, frequency_list):

    sampler = torch.distributions.normal.Normal(0, sigma)

    noise = sampler.sample(pred.shape).clamp(0, 1).to(pred.device)

    pred = pred + (noise.abs().permute(0, 2, 3, 1) * frequency_list / frequency_list.max()).permute(0, 3, 1, 2)

    loss = torch.nn.functional.cross_entropy(pred, target)

    return loss

\end{lstlisting}
\end{algorithm}

\section{Computational Overhead} \label{sec-sup:training-time}

We list parts of training time comparison in \cref{tab:training_time}, which suggests the computational overhead introduced by \method is limited and has a trivial impact on the overall training time.
As a plug-in design, \method demonstrates its superiority.

\vspace{-5pt}
\begin{table}[!ht]
\centering
\caption{
Training time comparison (with 8 V100 GPUs).
}
\label{tab:training_time}
\vspace{-10pt}
\setlength{\tabcolsep}{10pt}
\scalebox{0.85}{
\begin{tabular}{cc|cc}
\toprule
Backbone & Decoder & \textit{w/o} \method & \textit{w/} \method \\
\midrule
HRNet-18 & OCRHead & 20h11m & 21h07m (+4.6\%)\\
\midrule 
ResNet50&UperHead & 16h20m &16h47m (+2.8\%)\\
\bottomrule
\end{tabular}}
\vspace{-10pt}
\end{table}

\section{Estimation from the Labeled Data} \label{sec-sup:estimation-labeled}

Under semi-supervised settings, we have tried to estimate the distribution from the labeled data and found the overall performance improvement is limited. 
The results are presented in \cref{tab:estimation}.
We think this is due to the bias in estimating the full distribution from a small number of samples.

\vspace{-5pt}
\begin{table}[!ht]
\centering
\caption{
Experiments under semi-supervised settings. ST indicates self-training baseline, $\dag$ denotes estimation from the labeled data only, and $\ddag$ means \method estimation strategy described in the paper.
}
\label{tab:estimation}
\vspace{-10pt}
\setlength{\tabcolsep}{14pt}
\scalebox{0.85}{
\begin{tabular}{l|lll}
\toprule
Partition & ST & ST+BLV$^{\dag}$& ST+BLV$^{\ddag}$  \\
\midrule
1/16 & 68.21 & 68.22\up{0.01}& \textbf{69.26\up{1.05}} \\
\midrule
1/8 & 72.01 & 72.21\up{0.20} & \textbf{73.27\up{1.26}} \\
\bottomrule
\end{tabular}}
\vspace{-5pt}
\end{table}

\section{Compare with the GCL mothods} \label{sec-sup:compare with class}
Due to similar motivation with GCL~\cite{li2022long}, we add a detailed comparison on UDA Segmentation task in ~\cref{tab:sup_gcl}.
We also resample the training pixels to match the ``CBEN'' component in their paper.

\begin{table}[!ht]
\centering
\caption{
\textbf{Results on segmentation task.}
}
\setlength{\tabcolsep}{20pt}
\label{tab:sup_gcl}
\vspace{-8pt}
\begin{tabular}{ccc}
\toprule
 Baseline &  GCL &  \textbf{\method}  \\
\midrule
 55.9    &   56.1\up{0.2}  &   \textbf{59.0\up{3.1}}    \\
\bottomrule
\end{tabular}
\vspace{-10pt}
\end{table}
% Besides, we also implement our method for classification task with ResNet-50 backbone on ImageNet-LT dataset in ~\cref{tab:sup_gcl_cls}.

% \begin{table}[!ht]
% \centering
% \caption{
% %
% \textbf{Results on classification task.}
% %
% }
% \setlength{\tabcolsep}{18pt}
% \label{tab:sup_gcl_cls}
% \vspace{-8pt}
% \begin{tabular}{ccc}
% \toprule
%  Baseline &  \textbf{GCL} &  \method  \\
% \midrule
%   44.5   & \textbf{54.9\up{10.4}}    & 50.6\up{6.1}       \\
% \bottomrule
% \end{tabular}
% \vspace{-10pt}
% \end{table}
%
%These two tables 
This table indicates that \method is more suitable for segmentation task.